\PassOptionsToPackage{table}{xcolor}
\documentclass{article}
\usepackage{newunicodechar}
\newunicodechar{≈}{\approx}

\usepackage[authoryear, sort&compress, round]{natbib}
\usepackage[preprint,logo]{agibot_enerverse}

\usepackage[utf8]{inputenc} 
\usepackage[T1]{fontenc}    
\usepackage{hyperref}       
\usepackage{url}            
\usepackage{booktabs}       
\usepackage{amsfonts}       
\usepackage{nicefrac}       
\usepackage{microtype}      
\usepackage{xcolor}
\usepackage{amsmath}
\usepackage{graphicx}
\usepackage{caption}
\usepackage{subcaption}
\usepackage{wrapfig}
\usepackage{pifont}
\usepackage{colortbl}
\usepackage{multirow}
\usepackage[table]{xcolor}
\usepackage[absolute,overlay]{textpos}
\usepackage{placeins}

\definecolor{green}{RGB}{0,150,10}
\definecolor{blue}{RGB}{0,148,181}
\definecolor{orange}{RGB}{194,153,107}

\definecolor{grey}{HTML}{999999}
\definecolor{lightblue}{HTML}{B0C4DE}
\definecolor{purple}{HTML}{E3BBED}
\definecolor{orange}{HTML}{ffdab9}
\definecolor{cadetblue}{HTML}{5F9EA0}
\definecolor{darksalmon}{rgb}{0.91, 0.59, 0.48}
\definecolor{forestgreen}{rgb}{0.13, 0.55, 0.13}
\definecolor{BlueGreen}{rgb}{0.0, 0.55, 0.55}
\definecolor{RedOrange}{rgb}{1.0, 0.27, 0.0}

\usepackage{xspace}

\newcommand{\pizerofive}{$\pi_{0.5}$\xspace}
\newcommand{\Aprime}{$A'$\xspace}
\newcommand{\Bprime}{$B'$\xspace}

\colorlet{maniorange}{reportorange}
\definecolor{takeawaybg}{HTML}{FFFAF4}
\definecolor{takeawaytitlebg}{HTML}{FDEAD7}
\newtcolorbox{takeawaybox}[1][]{%
  enhanced, breakable, colback=takeawaybg, colframe=reportline,
  before skip=14pt,
  boxrule=0.5pt, arc=2mm, left=8pt, right=8pt, top=6pt, bottom=6pt,
  fonttitle=\sffamily\bfseries, coltitle=reportorange,
  colbacktitle=takeawaytitlebg, #1}
\newcommand{\takeawayhead}[1]{\par\smallskip{\bfseries\color{maniorange}#1}\par}

\title{Hidden Variables Behind Real-Robot Data Quality:\\
Lessons from Hundreds of Thousands of Hours}

\author{%
  {\normalfont\normalsize Maniformer Real-World Manipulation Team}\\[0.5em]
  {\normalfont\normalsize November 11, 2026}\\[0.5em]
  {\normalfont\normalsize\href{https://teleop.agibot.com/}{\textcolor{reportorange}{https://teleop.agibot.com/}}}
}

\begin{document}

\maketitle

\vspace{-2.2em}
\begin{center}
  \includegraphics[width=\linewidth]{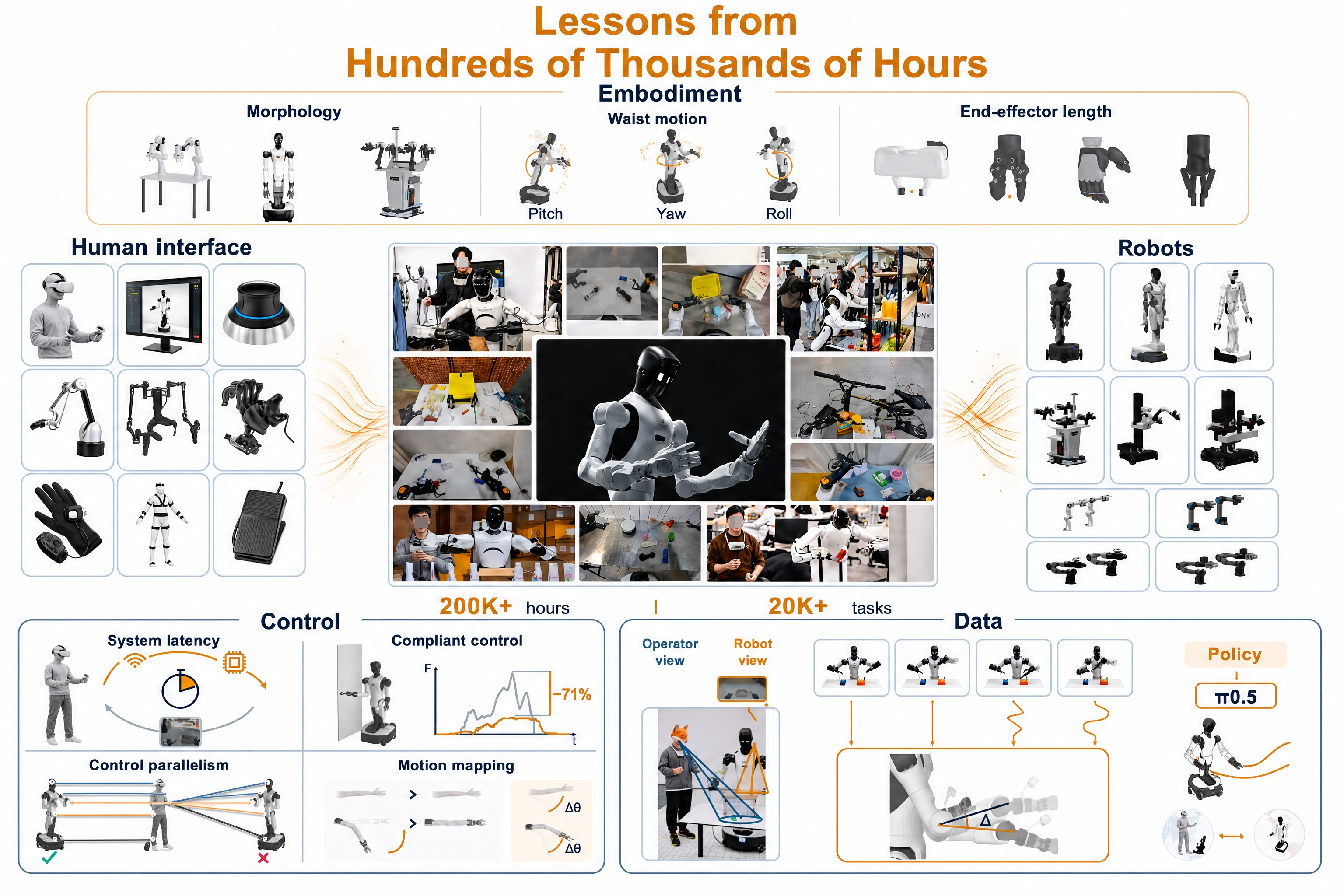}
  \captionof{figure}{\textbf{Overview.} Operators collect demonstrations through a range of human interfaces (left); the resulting data (center; 200K+ collection hours across 20K+ tasks) train policies across robot embodiments (right). We study three factors that shape these data: embodiment (\S\ref{sec:embodiment}), teleoperation control (\S\ref{sec:control}), and data quality (\S\ref{sec:data}).}
  \label{fig:teaser}
\end{center}

\begin{abstract}
As real-robot data collection scales, collection-system design shapes both operator efficiency and the behaviors learned by robot policies. Workspace limitations, control-interface friction, and latency can introduce pauses, detours, and corrective motions into otherwise successful demonstrations. We investigate these effects across embodiment, teleoperation control, and data consistency, combining a kinematic study of 25 dual-arm robots, controlled teleoperation experiments, and real-robot evaluation of fine-tuned \pizerofive policies. The embodiment study shows that positional reachability overstates orientation-aware task coverage and that modest shoulder motion can substantially expand the effective workspace. Teleoperation experiments show that parallelizing body-part control reduces task completion time by 43--53\%, while compliant control improves success and reduces peak contact forces on constrained and contact-rich tasks. Policy-learning experiments show that demonstrations must capture the visual information underlying action choices: active-search demonstrations improve arm-selection accuracy from chance (50\%) to 100\%. Calibration experiments reveal that mixed-calibration training data remain usable in our tested configurations when an inference-aligned subset is retained, whereas inference-time miscalibration causes all 30 trials to fail. Policies also retain approximately 90\% of the extra path length in detour-contaminated demonstrations, showing how inefficient operator motions persist in learned behavior. Together, these findings connect collection-system design to demonstration quality and downstream policy execution. They motivate evaluating effective workspace, control efficiency, observation completeness, calibration alignment, and motion quality before scaling collection, and addressing system-induced compensations at their source.
\end{abstract}

\section{Introduction}
\label{sec:intro}

As real-robot data collection continues to scale, the design of the collection system becomes increasingly consequential for both collection efficiency and demonstration quality. The robot's embodiment determines which task poses are feasible, the teleoperation interface shapes how operators execute motions, and the sensing and calibration setup determines how those motions are represented in the recorded data. These system choices influence both the effort required to collect demonstrations and the behaviors that robot policies can learn from them. Understanding their effects is therefore an essential part of building a reliable data collection pipeline.

Consider an ordinary grasping task. The robot may respond quickly to commands, yet the video presented to the operator can lag behind its actual motion. The operator notices an overshoot only after it has occurred, pauses to wait for feedback, and then corrects the trajectory. Limited workspace or frequent control-mode switching can introduce further adjustments. The grasp may ultimately succeed, while the recorded demonstration includes the pauses, detours, and corrections needed to compensate for the collection system. \textbf{Task success alone does not reveal the quality of the resulting demonstration.} Some compensatory motions may subsequently be reproduced by a policy trained on those trajectories.

Other data issues arise from the information and coordinate relationships available during collection. An operator may choose an action using visual cues outside the robot's recorded camera views, leaving the policy without the evidence needed to make the same decision. Differences in joint-zero calibration can cause the same recorded joint values to correspond to different physical configurations across robots or collection sessions. These issues affect different aspects of the learning problem: whether observations support action choices, whether recorded states and actions are consistent, and whether demonstrated motions are efficient. \textbf{Demonstration quality therefore requires examining the relationship between the collection setup, the recorded signals, and the resulting policy behavior.}

We investigate these issues through three complementary perspectives (Figure~\ref{fig:teaser}):
\begin{itemize}
  \item \textbf{Embodiment (\S\ref{sec:embodiment}):} Can the robot perform the task in the required workspace? Using kinematic models of 25 dual-arm robots, we examine how arm morphology, waist motion, and end-effector length affect orientation-aware workspace coverage and joint-limit margin.
  \item \textbf{Teleoperation control (\S\ref{sec:control}):} Can the operator execute the task smoothly and efficiently? Through representative tasks and controlled experiments, we study control parallelism, motion mapping, compliant control, system latency, and peripheral configuration, examining their effects on task completion, operator efficiency, and interaction quality.
  \item \textbf{Data quality (\S\ref{sec:data}):} Are demonstrations informative, consistent, and efficient? Through fine-tuning and real-robot evaluation of \pizerofive policies~\citep{black2025pi05}, we examine how missing visual decision cues, joint-zero calibration mismatch, and redundant demonstrated motions affect learned behavior.
\end{itemize}

These studies provide empirical evidence under representative tasks and configurations, together with practical recommendations for embodiment selection, teleoperation design, and data acceptance. \textbf{Our central perspective is that collection-system design and demonstration quality should be evaluated together: system limitations can shape the trajectories that enter training, while incomplete observations and inconsistent calibration can change what those trajectories teach.} For practitioners scaling real-robot data collection, this motivates checking task feasibility, operator control, and the recorded learning signals before expanding collection.

\begin{takeawaybox}[title={Key Takeaways}]
\takeawayhead{Embodiment: can the robot do it? (\S\ref{sec:embodiment})}
\begin{itemize}[leftmargin=1.2em, label=\textcolor{maniorange}{\textbullet}]
  \item \textbf{Select embodiments by orientation-aware effective workspace, not positional reach.} Reach overstates suitability: humanoid and wheeled-lift arms reach 55\% and 53\% of a sagittal slice through the shoulder but cover only 41\% and 33\% of tool orientations, against 91\% and 75\% for table-mounted arms.
  \item \textbf{For tabletop work, non-humanoid layouts have the advantage.} At their best table heights, table-mounted arms reach 80\% downward coverage and wheeled-lift arms 68\% forward coverage, against 52\%/58\% for the humanoid.
  \item \textbf{Treat a small amount of waist motion as the cheapest workspace multiplier, and prioritize pitch and yaw over roll.} Allowing 15\,cm of shoulder translation enlarges the reachable cross-section 3.4--5.5$\times$; in human motion data, pitch and yaw move 1.2--1.9$\times$ as much as roll.
  \item \textbf{Do not buy reach with tool length.} A 10\,cm longer tool adds 39\% reachable positions but pushes solutions toward joint limits; weigh coverage and joint margin equally, and extend the workspace through the waist first.
\end{itemize}
\takeawayhead{Control system: can the operator control it smoothly? (\S\ref{sec:control})}
\begin{itemize}[leftmargin=1.2em, label=\textcolor{maniorange}{\textbullet}]
  \item \textbf{Parallelize control and eliminate high-frequency mode switching.} Moving the waist to IK and the base and head to the headset cut completion time by 43--53\%; train operators to overlap motions (2--4.5$\times$ throughput gains over 5--7 weeks).
  \item \textbf{Choose the motion mapping by task range and continuity.} Prefer absolute mapping for large-range tasks that require continuous trajectories; prefer relative mapping with clutching for limited ranges or frequent operator repositioning.
  \item \textbf{Provide compliance for mechanism-constrained, continuous-contact, and bimanual closed-chain tasks.} Force control reached 100\% success against 80--87\% for position control and cut peak contact force by 55--71\%; such tasks make up 18.5\% of 207k collection hours, and a further 38.7\% would benefit. Stiff position control remains adequate for low-contact pick-and-place.
  \item \textbf{Optimize the measured human-in-the-loop latency, not algorithm runtime.} Sub-millisecond filters can add over 200\,ms of phase lag; the slowest required body part (typically gripper, waist, or base) sets the loop. Aim for latency below about 100\,ms; further reductions yield diminishing returns.
  \item \textbf{Match peripherals to the task and to shift length.} Use screens for long fixed-station sessions (about 62\,min before discomfort versus 19\,min in VR); use VR with body tracking for active viewpoint control and whole-body tasks; use a SpaceMouse or force-feedback exoskeleton for fine or contact-sensitive work.
\end{itemize}
\takeawayhead{Data quality: can the data explain the action? (\S\ref{sec:data})}
\begin{itemize}[leftmargin=1.2em, label=\textcolor{maniorange}{\textbullet}]
  \item \textbf{Make sure the recorded observations explain every decision.} For occlusions, out-of-view targets, and branching decisions, collect beyond line of sight (BLOS) and make active observation part of the demonstration: arm-choice accuracy rose from chance (50\%) to 100\%.
  \item \textbf{Make collection--inference calibration alignment a pre-deployment gate.} Record calibration versions, verify joint readings at a reference pose, and compensate per-joint zero offsets; a miscalibrated inference robot failed in all 30 trials. Mixed-calibration history need not be discarded, but keep an inference-aligned anchor subset (66.7\% vs.\ 26.7\% success with vs.\ without it); zero-mean offsets do not cancel out (20\%).
  \item \textbf{Keep actions clean, and accept demonstrations on path efficiency, continuity, and duration, not success alone.} Policies retain about 90\% of the extra path length but only 45\% of the extra jerk; 80\% detour data slowed execution by 29\%. Fix system-induced corrections at the source, and keep purposeful information-gathering motions.
\end{itemize}
\takeawayhead{Overall}
\textbf{Before scaling collection, validate the embodiment's effective workspace, operator efficiency, and data learnability on representative tasks.}
\end{takeawaybox}

\section{Embodiment Design: Can the Robot Carry Out the Operator's Intent?}
\label{sec:embodiment}

Embodiment design should first match the manipulation requirements of the target tasks. That an arm can reach a target position does not mean it can operate there with the orientation the task requires, or that it retains sufficient joint range after arriving. Even when a target pose is feasible, if the required configuration lies close to joint limits, or the body posture must be adjusted frequently to cover the work area, the operator's burden increases and extra adjustment motions enter the demonstration. \textbf{Reachability is strongly constrained by the end-effector orientation the task requires, so the effective workspace is smaller than the reachable workspace. Degrees of freedom and morphology should therefore be chosen according to the workspace in which the task is best performed.}

\subsection{Morphology Should Match the Task's Operating Directions}
\label{sec:morphology}

Different tasks impose different requirements on manipulation position and end-effector orientation. In tabletop grasping, placing, and wiping, for example, the end effector typically approaches the target from the front or from above. Using URDF models of 25 mainstream humanoid and non-humanoid dual-arm robots, we test multiple tool-axis directions at each of 10{,}075 positions in a sampled volume and use inverse kinematics (IK) to compute the fraction of feasible directions at each position.

\begin{figure}[tbp]
  \centering
  \includegraphics[width=\linewidth]{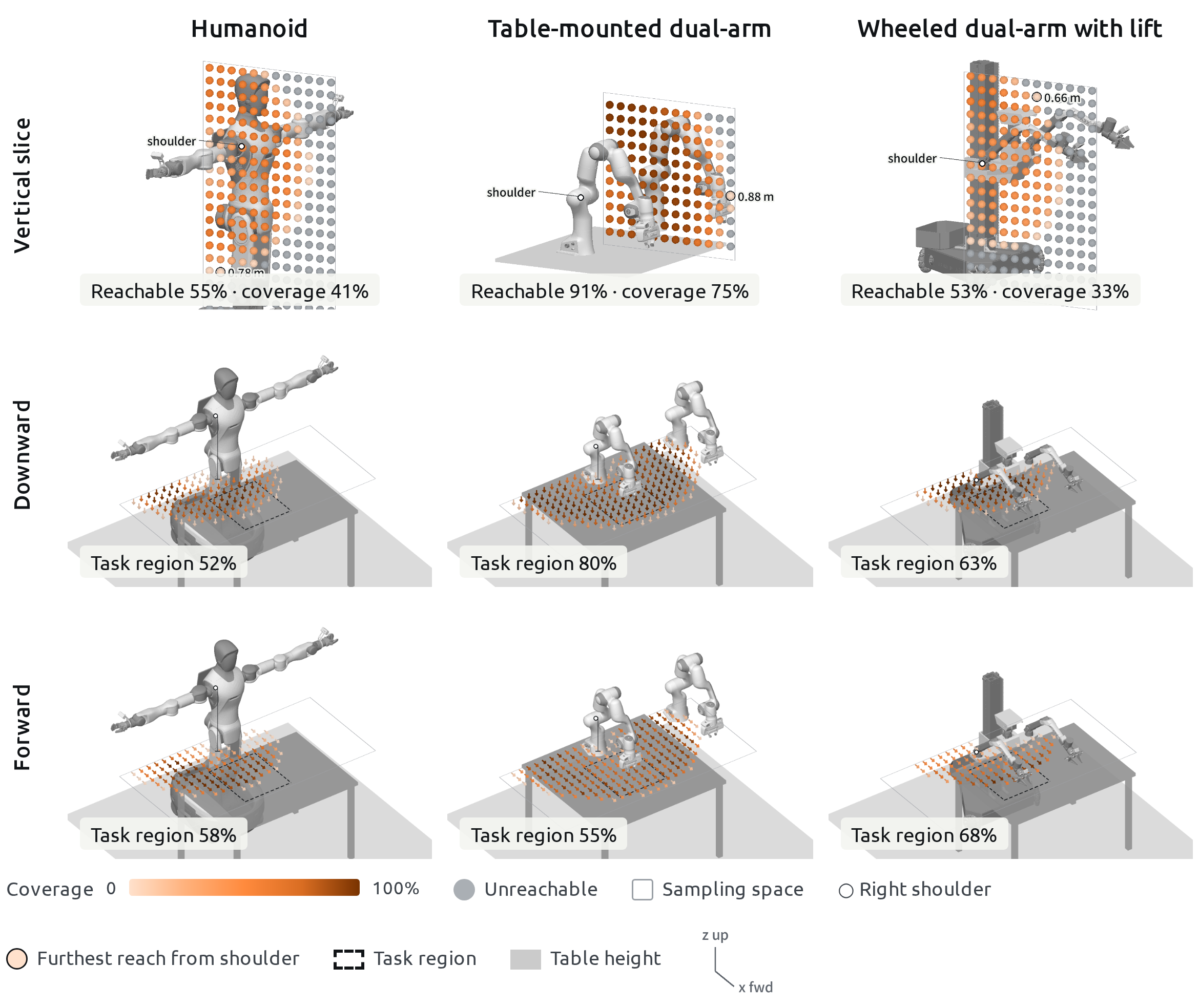}
  \caption{\textbf{Orientation coverage of three morphologies.} Top row: right-arm reachability and coverage in a vertical slice through the right shoulder; each position is tested against 64 discretized tool-axis directions, color shows the fraction that is solvable, and gray dots are unreachable (a position counts as reachable if at least one direction is solvable). Middle and bottom rows: coverage of the downward subset (9 directions within 45$^\circ$ of gravity) and the forward subset (11 directions within 45$^\circ$ of $+X$) on a horizontal slab just above the table, each morphology at its optimal table height; the dashed rectangle marks the task region (table footprint, 5--35\,cm high).}
  \label{fig:coverage}
\end{figure}

\textbf{Positional reachability does not imply that the required end-effector orientations are feasible, and so cannot by itself guide the choice of morphology.} The top row of Figure~\ref{fig:coverage} shows reachability and orientation coverage for the right arm of three morphologies within a $0.9\,\text{m}\times1.8\,\text{m}\times1.8\,\text{m}$ volume, with 64 discretized tool-axis directions tested at each position. Reachability only requires that at least one direction is solvable, whereas coverage reflects how many tool-axis directions each position can satisfy. In the sagittal slice through the right shoulder, \textbf{in the best case the humanoid and the wheeled dual-arm with lift have similar reachability (55\% and 53\%) and mean coverage (41\% and 33\%), while the table-mounted dual-arm is markedly higher on both (91\% and 75\%)}. Comparing arms by positional reach alone easily overestimates their suitability for a specific manipulation task.

Once the end-effector orientations required by tabletop tasks are taken into account, the effective workspaces of the different morphologies diverge (Figure~\ref{fig:coverage}, middle and bottom rows). Centered on gravity and on the robot's forward direction ($+X$), we select 9 downward and 11 forward directions within a 45$^\circ$ cone and maximize the mean coverage of each set over table height. At their respective optimal table heights, the table-mounted dual-arm attains the highest downward coverage (80\%) with 55\% forward coverage; the wheeled dual-arm with lift attains the highest forward coverage (68\%) with 63\% downward; and the humanoid attains 52\% and 58\%, respectively. \textbf{For tabletop manipulation, non-humanoid morphologies hold an advantage over humanoids.}

\subsection{Limited Waist Motion Substantially Enlarges the Tabletop Workspace}
\label{sec:waist}

Can the workspace limitation of humanoid morphologies on tabletop tasks be compensated for by waist motion? Waist motion changes the position and orientation of the shoulder, thereby extending the arm's effective workspace. To quantify this effect, we construct representative humanoid and non-humanoid arms by averaging the link parameters of each class and compare their workspaces when the shoulder is allowed to translate.

\begin{figure}[tbp]
  \centering
  \includegraphics[width=\linewidth]{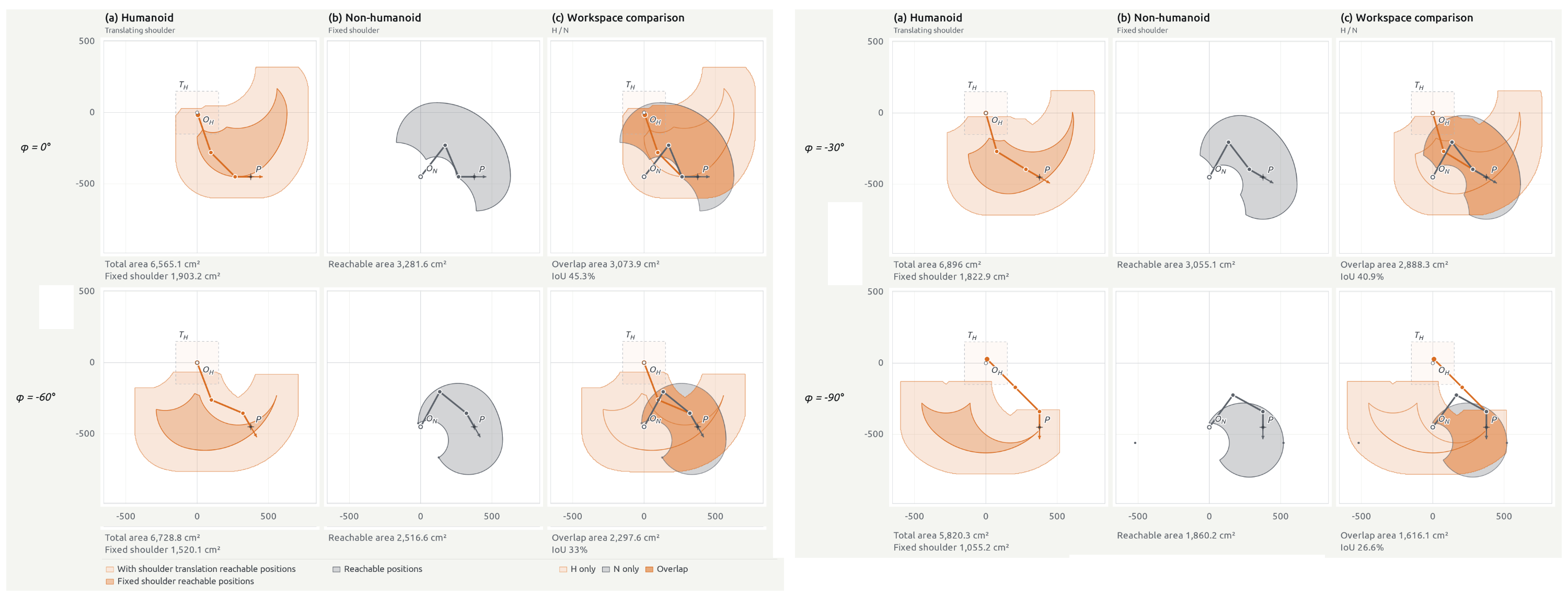}
  \caption{\textbf{Effect of shoulder translation on humanoid and non-humanoid workspaces.} Representative link lengths and joint limits for each class are derived from 25 mainstream robot URDFs, and planar reachable cross-sections are compared at four fixed end-effector orientations $\varphi$. Within each orientation: (a) humanoid, where dark orange is the reachable region with a fixed shoulder and light orange the extension when the shoulder may translate horizontally and vertically; (b) non-humanoid; (c) overlap of the two classes. Axes in mm.}
  \label{fig:workspace_matrix}
\end{figure}

Figure~\ref{fig:workspace_matrix} shows reachable cross-sections at four end-effector orientations (0$^\circ$, $-30^\circ$, $-60^\circ$, $-90^\circ$). \textbf{Allowing the shoulder to translate within a sphere of radius 15\,cm enlarges the reachable cross-sectional area to 3.4--5.5$\times$ that of a fixed shoulder}, substantially relieving the humanoid's tabletop limitation. \textbf{In a side-by-side comparison, the humanoid's reachable cross-section with shoulder translation can be up to 3$\times$ that of the non-humanoid.}

The individual waist degrees of freedom also contribute differently. We analyze 5{,}496 human motion sequences from the AMASS dataset~\citep{mahmood2019amass} together with 15 hours of whole-body mobile tabletop manipulation captured with PICO trackers. In both datasets, the geometric means of the pitch/roll and yaw/roll ratios of torso-relative-to-pelvis rotation exceed 1 (AMASS: 1.61 and 1.88; PICO: 1.52 and 1.21; Table~\ref{tab:waist_ratio}), indicating that pitch and yaw carry larger relative motion. This agrees with their kinematic roles---pitch repositions the shoulder and yaw changes the arm's approach direction---whereas the function of roll can largely be substituted by base or leg motion. \textbf{System design should therefore evaluate the waist's ability to dynamically reposition the shoulder, rather than relying solely on changing arm morphology or link lengths. Where waist DoFs must be limited, pitch and yaw should be prioritized.}

\begin{table}[!htb]\centering
  \centering
  \caption{\textbf{Waist rotation relative to roll in human motion data.} Geometric means of the pitch/roll and yaw/roll ratios of spine3-relative-to-pelvis rotation in a forward--left--up frame, with grouped-bootstrap 95\% confidence intervals.}
  \label{tab:waist_ratio}
  \begin{tabular}{lcc}
    \toprule
    Dataset & Pitch / roll & Yaw / roll \\
    \midrule
    AMASS~\citep{mahmood2019amass} & 1.61 [1.51, 1.70] & 1.88 [1.72, 2.05] \\
    PICO motion capture & 1.52 [1.22, 1.87] & 1.21 [1.04, 1.43] \\
    \bottomrule
  \end{tabular}
\end{table}

\subsection{End-Effector Length Trades Coverage Against Local Joint Margin}
\label{sec:tip}

End-effector length changes the geometric relation between the wrist and the actual point of operation, and hence the joint configuration required for the same task pose. A longer end effector can extend positional coverage but may also push the joints closer to their limits, so its length must balance reach against joint margin.

\begin{figure}[tbp]
  \centering
  \includegraphics[width=\linewidth]{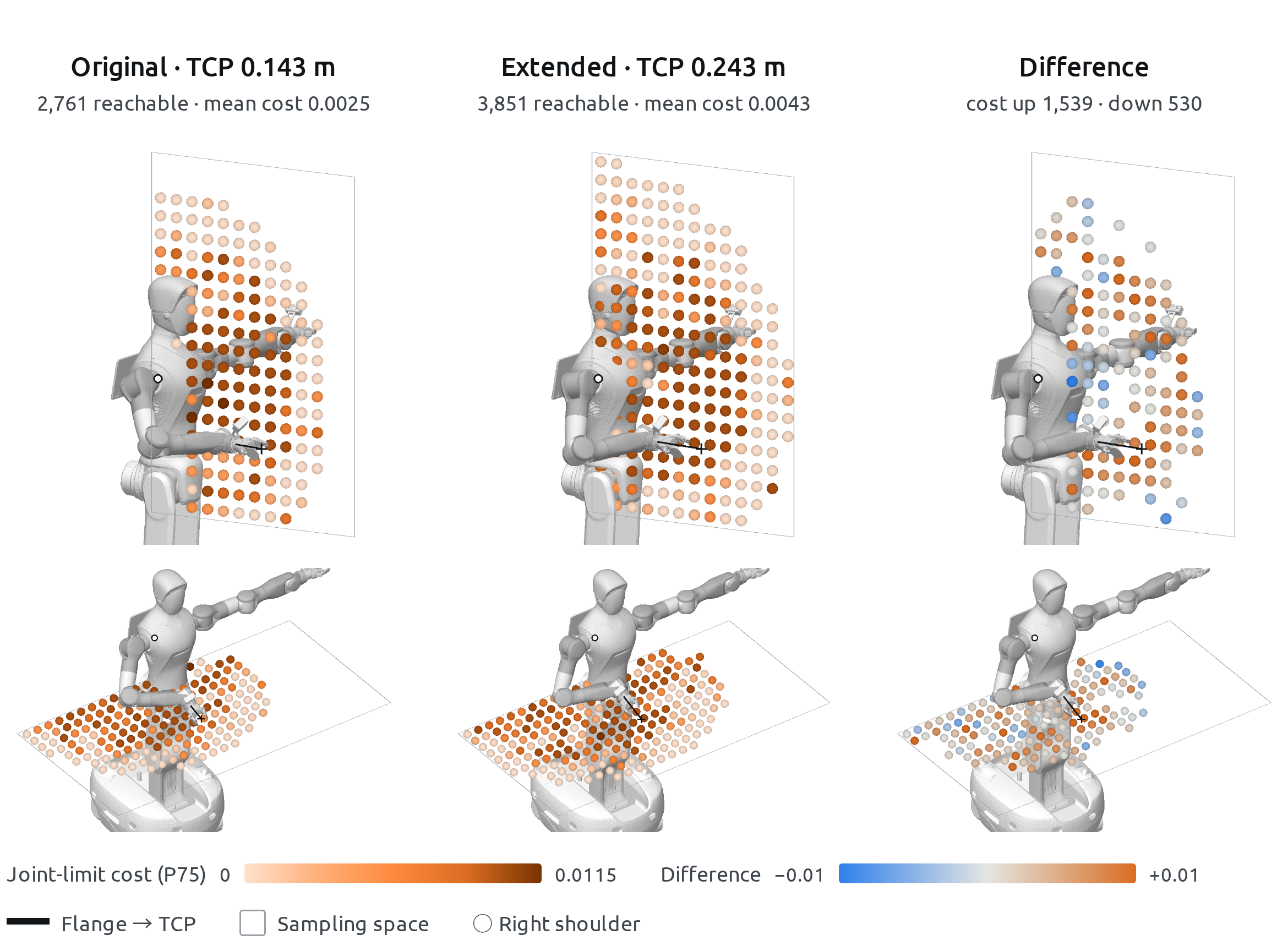}
  \caption{\textbf{Effect of end-effector length on coverage and joint-limit cost.} Left and middle: TCP distances of 0.14308\,m and 0.24308\,m over the same 10{,}075 sampled positions. Targets constrain TCP position and tool-axis direction, with free rotation about the tool axis. Color shows the 75th percentile (P75) of joint-limit cost over successful directions (shared color scale; blue is low, red is high). Right: P75(extended)\,$-$\,P75(original) over directions solvable in both; positive values indicate higher cost after extension.}
  \label{fig:tip_length}
\end{figure}

We increase the wrist-to-TCP distance from about 14.31\,cm to 24.31\,cm and compare the two configurations over the same 10{,}075 sampled positions. The number of positions with at least one solvable direction rises from 2{,}761 to 3{,}851, an increase of about 39.5\%. However, over the target directions solvable in both configurations, the IK solutions of the longer tool exhibit higher joint-limit cost (Figure~\ref{fig:tip_length}). This not only reduces dexterity at the target~\citep{yoshikawa1985manipulability} but also raises the risk of approaching kinematic singularities and weakens dynamic stability during operation.

\textbf{When choosing tool length, task coverage and joint margin should be optimized jointly and weighted equally.} For tasks that require frequent fine adjustment, avoid unnecessary tool extension once grasp size and approach requirements are met, so as to preserve joint margin.

Overall, embodiment design should coordinate arm selection, waist DoFs, and end-effector length according to the positions and orientations the task requires. \textbf{When the morphology struggles to meet task requirements, extend the effective workspace through the waist first, rather than trading tool length for reach.} Sound embodiment design determines not only whether a task can be completed, but also how many extra adjustments the operator must make. Whether operator intent can be mapped effectively onto this effective workspace---making full use of the embodiment's motion capability---is the responsibility of the control system.

\section{Control System: Can the Operator Convey Intent to the Robot Smoothly?}
\label{sec:control}

The control system is the bridge between human intent and robot motion. \textbf{An ideal control system should map the operator's motion intent accurately into executable target trajectories that fully exploit the embodiment, while minimizing the extra cognitive and physical load the operator incurs in adapting to the interface.} Frequent mode switching, unintuitive motion mapping, repeated corrections caused by contact errors, and latency and peripheral limitations can all lengthen demonstrations and introduce unnecessary pauses and adjustments into trajectories. We evaluate control systems along three axes: operating efficiency, ease of learning, and task success.

\subsection{Control Parallelism: Reducing Serial Operation and Mode Switching}
\label{sec:parallelism}

Whole-body control (WBC) coordinates the mobile base, waist, head, both arms, and end effectors simultaneously. Operating all joints and subsystems in parallel significantly improves efficiency. Parallelism alone, however, is not sufficient; an intuitive human--robot kinematic mapping is equally important. Conversely, if multiple subsystems share one controller and rely on mode switching, the operator must perform base motion, waist pitch, viewpoint adjustment, and arm reaching serially. This lengthens operation and readily introduces discontinuous pauses into demonstrations, reducing data quality.

\begin{figure}[tbp]
  \centering
  \includegraphics[width=\linewidth]{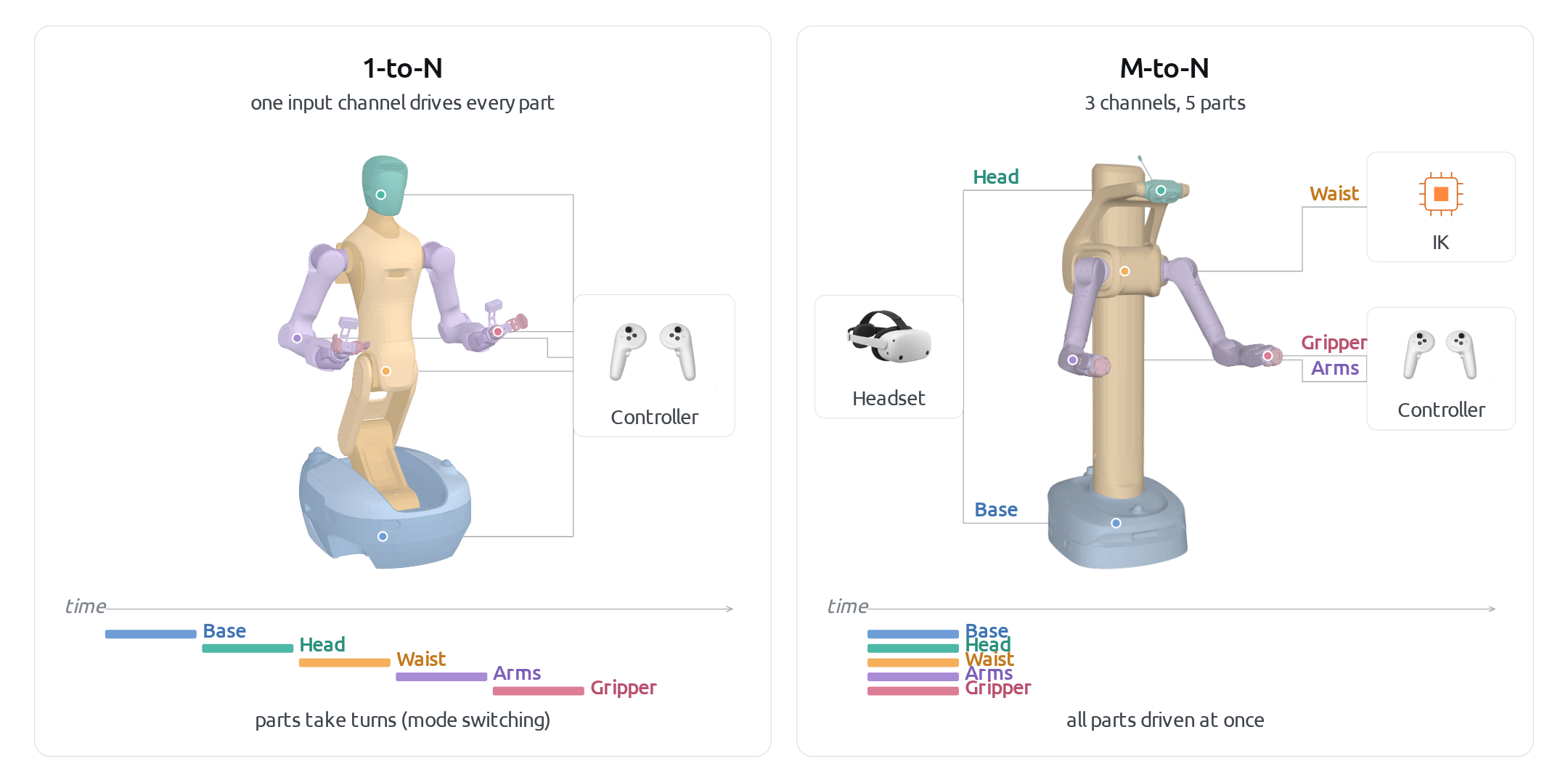}
  \caption{\textbf{Input-channel mapping.} Left (1-to-N): a single input channel drives all body parts, which take turns via mode switching. Right (M-to-N): three channels drive five parts---headset for base and head, IK for the waist, and controllers for arms and grippers---so all parts can move at once.}
  \label{fig:input_mapping}
\end{figure}

In the 1-to-N configuration (Figure~\ref{fig:input_mapping}, left), the controller time-multiplexes different body parts via mode switching. The M-to-N configuration distributes control responsibilities across separate input channels and automatic modules: the headset drives the base and head, IK coordinates the waist from the hand and head targets, and the controllers drive the arms and grippers. This lets the operator adjust body posture and viewpoint while approaching the target, reducing the burden of serial operation and mode switching.

\begin{figure}[tbp]
  \centering
  \includegraphics[width=\linewidth]{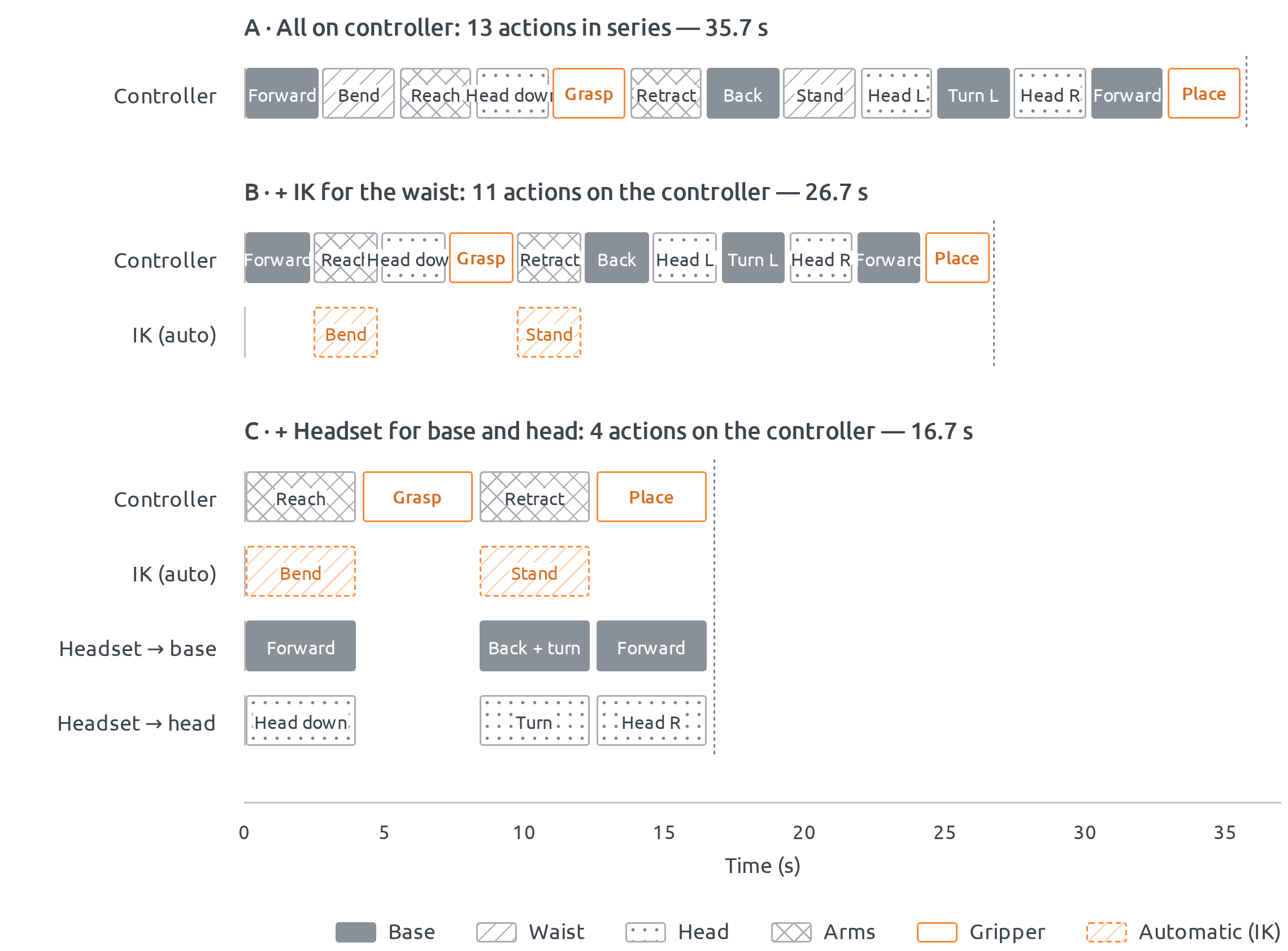}
  \caption{\textbf{Action sequences under the three configurations} for the floor-pick task (configurations as in Table~\ref{tab:parallel_results}). Colors distinguish base, waist, head, arms, and end effectors; hatching marks actions solved automatically by IK. Total durations are measured medians over 30 trials; action order comes from task logs, while individual segment lengths are evenly allocated for illustration and do not estimate per-action durations.}
  \label{fig:parallel_sequence}
\end{figure}

We compare three configurations on two mobile manipulation tasks, ``pick an item from the floor and place it on a table behind the robot'' and ``wipe two separated regions of a table'': in A, every body part is controlled by the hand controllers; B adds IK control of the waist to A; C additionally hands the base and head to the headset. Each configuration was run 30 times per task. Median completion time for the floor-pick task fell from 35.7\,s to 26.7\,s and 16.7\,s, i.e., 25\% and 53\% shorter than A; for wiping it fell from 27.0\,s to 19.3\,s and 15.3\,s, i.e., 28.5\% and 43\% shorter (Table~\ref{tab:parallel_results} and Figure~\ref{fig:parallel_sequence}). \textbf{As control responsibilities are progressively decoupled and more body parts can be operated in parallel, task efficiency improves markedly.}

\begin{table}[!htb]\centering
  \centering
  \caption{\textbf{Completion time after decoupling control of body parts}, median of 30 trials per configuration (s); parentheses give the reduction relative to A. A: every body part on the hand controllers; B: A plus IK control of the waist; C: B plus headset control of base and head.}
  \label{tab:parallel_results}
  \begin{tabular}{lccc}
    \toprule
    Task & A & B & C \\
    \midrule
    Floor pick $\rightarrow$ table behind & 35.7 & 26.7 ($-$25\%) & 16.7 ($-$53\%) \\
    Wipe two spots on a table & 27.0 & 19.3 ($-$28.5\%) & 15.3 ($-$43\%) \\
    \bottomrule
  \end{tabular}
\end{table}

The benefit of parallel control also depends on operator proficiency. Our collection logs show that novices tend to operate body parts one at a time, while experienced operators overlap motions more often. In 2D-screen collection, for example, six operators collecting continuously improved from about 180--400 to 790--1{,}030 successful items per effective hour over 5--7 weeks, roughly 2--4.5$\times$ their initial level. \textbf{Whole-body teleoperation systems should first eliminate high-frequency mode switching and unnecessary serial operation, and help operators exploit parallel control through intuitive mappings and training.}

\subsection{Motion Mapping: Balancing Continuity and Flexibility}
\label{sec:mapping}

Motion mapping determines how the operator's movements are converted into robot motion targets; its intuitiveness affects efficiency, learning cost, and operator preference. \emph{Absolute mapping} maps the operator's current hand or body pose to a robot target pose through a coordinate transform and scaling. Under fixed calibration and mapping parameters, the same human pose always corresponds to the same robot target, so spatial relationships are intuitive and no re-referencing is needed, which favors motion continuity. However, its operating range is bounded by the human workspace and the mapping scale, and large-range operation usually requires coordinated waist or base motion.

\begin{figure}[tbp]
  \centering
  \begin{subfigure}{\linewidth}
    \centering
    \includegraphics[width=\linewidth]{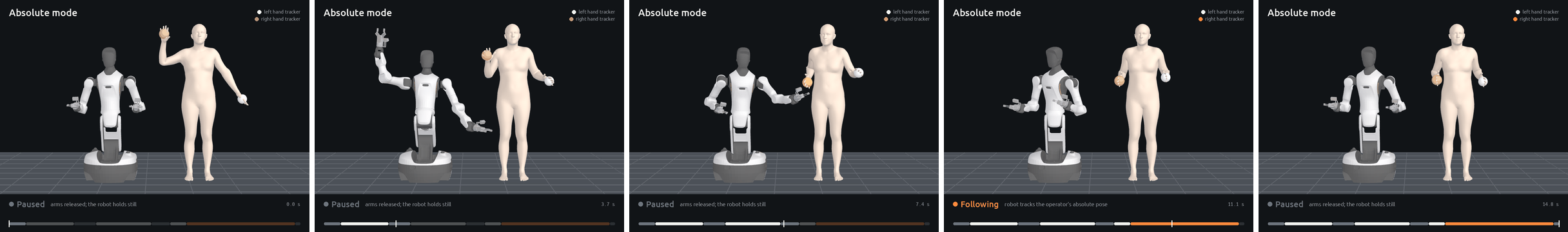}
    \caption{Absolute mapping}
  \end{subfigure}\\[4pt]
  \begin{subfigure}{\linewidth}
    \centering
    \includegraphics[width=\linewidth]{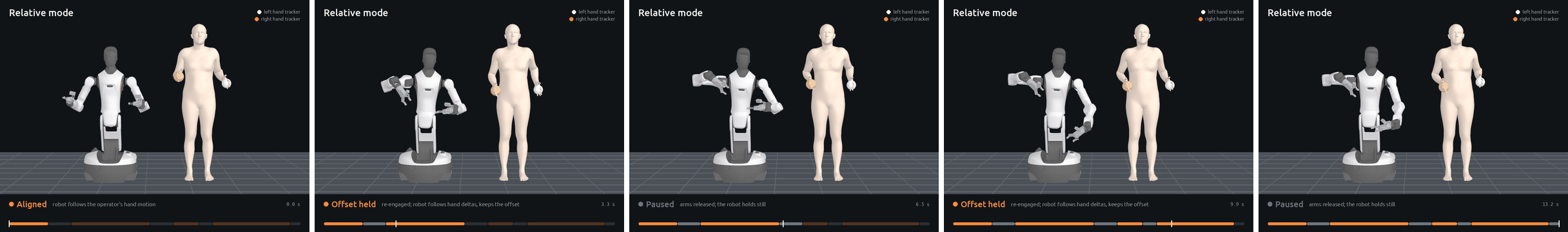}
    \caption{Relative mapping}
  \end{subfigure}
  \caption{\textbf{Replays of representative demonstrations under absolute and relative mapping}, shown as frames sampled at equal intervals from the operator's tracked body motion.}
  \label{fig:mapping_frames}
\end{figure}

\emph{Relative mapping} converts increments of controller motion into increments of the robot target pose. Through clutching, the operator can reposition themselves and keep controlling the robot within a limited human workspace. This suits scenarios that require frequent changes of operator posture, but clutching and control-state switching introduce extra pauses that break trajectory continuity (Figure~\ref{fig:mapping_frames}).

In large-scale collection, the choice of mapping is also shaped by operator habits, wearable burden, and sustained-operation needs. Line-of-sight (LOS) collection, for example, usually lets operators observe the scene directly and reposition freely, so relative mapping is highly practical for long continuous shifts. Absolute mapping maintains a stable spatial correspondence, but its convenience depends on the body-tracking range, control devices, and environment.

The motion mapping should be chosen according to task range, continuity requirements, and the operator's sustained working conditions. \textbf{For large-range tasks that emphasize trajectory continuity, absolute mapping is preferable; for limited-range operation or scenarios that require frequent operator repositioning, relative mapping offers greater flexibility.}

\subsection{Compliant Control: Reducing Corrections Caused by Contact Errors}
\label{sec:compliance}

Contact tasks require the robot to accommodate both the target trajectory and environmental constraints. When opening a door, the end effector is constrained by the hinge trajectory; when wiping, contact must be maintained continuously; and when carrying a rigid object with two hands, the two end effectors form a closed kinematic chain. In these tasks, high-stiffness position control tries to track the commanded target strictly and may generate large contact or internal forces when the target is inconsistent with the true constraint, forcing the operator to adjust the target repeatedly. \textbf{Mechanism-constrained, contact-rich, and bimanual closed-chain tasks place higher demands on compliance; their efficiency and success rate are strongly affected by it.}

Compliant control lets the robot adjust its motion response to contact, reducing reliance on precise trajectory tracking. Common approaches include impedance control, which regulates the force response to motion deviation through equivalent stiffness and damping~\citep{hogan1985impedance}, and admittance control, which adjusts the motion target according to measured external force, allowing the robot to accommodate environmental constraints within a certain range.

The need for compliant control in real collection is widespread. Our records contain about 20{,}253 task entries totaling 207{,}030 collection hours. By collection time, 18.5\% of the data clearly involves mechanism-constrained, continuous-contact, or bimanual closed-chain tasks that require compliant control, and a further 38.7\% corresponds to tasks such as smoothing, pressing, folding, tensioning, and tight fits that would benefit from it (Table~\ref{tab:compliance_share}).
To verify the role of compliant control in practice, we select four tasks---block stacking, door opening, whiteboard wiping, and bimanual box carrying---each performed by an experienced teleoperator under force and position control, 30 trials per configuration. Force control achieved 100\% success on all four tasks; position control also reached 100\% on block stacking, but only 83.3\%, 80\%, and 86.7\% on door opening, whiteboard wiping, and box carrying, respectively (Table~\ref{tab:force_vs_position}). \textbf{Compared with low-contact pick-and-place, compliant control shows a clearer success-rate advantage in mechanism-constrained, continuous-contact, and bimanual closed-chain tasks.}

\begin{table}[!htb]\centering
  \centering
  \caption{\textbf{Success rate (\%) by task and control mode}, 30 trials per configuration, and peak contact force (N) on representative trajectories.}
  \label{tab:force_vs_position}
  \begin{tabular}{llcccc}
    \toprule
    & & \multicolumn{2}{c}{Success (\%)} & \multicolumn{2}{c}{Peak force (N)} \\
    \cmidrule(lr){3-4}\cmidrule(lr){5-6}
    Task & Contact type & Force & Position & Force & Position \\
    \midrule
    Block stacking & Pick and place & 100 & 100 & 12.1 & 11.2 \\
    Door opening & Mechanism-constrained & 100 & 83.3 & 34.6 & 120.5 \\
    Whiteboard wiping & Contact-rich & 100 & 80.0 & 29.2 & 94.4 \\
    Box carrying (L / R hand) & Bimanual closed chain & 100 & 86.7 & 32.7 / 25.7 & 72.6 / 85.3 \\
    \bottomrule
  \end{tabular}
\end{table}

\begin{figure}[tbp]
  \centering
  \begin{subfigure}{0.49\linewidth}
    \includegraphics[width=\linewidth]{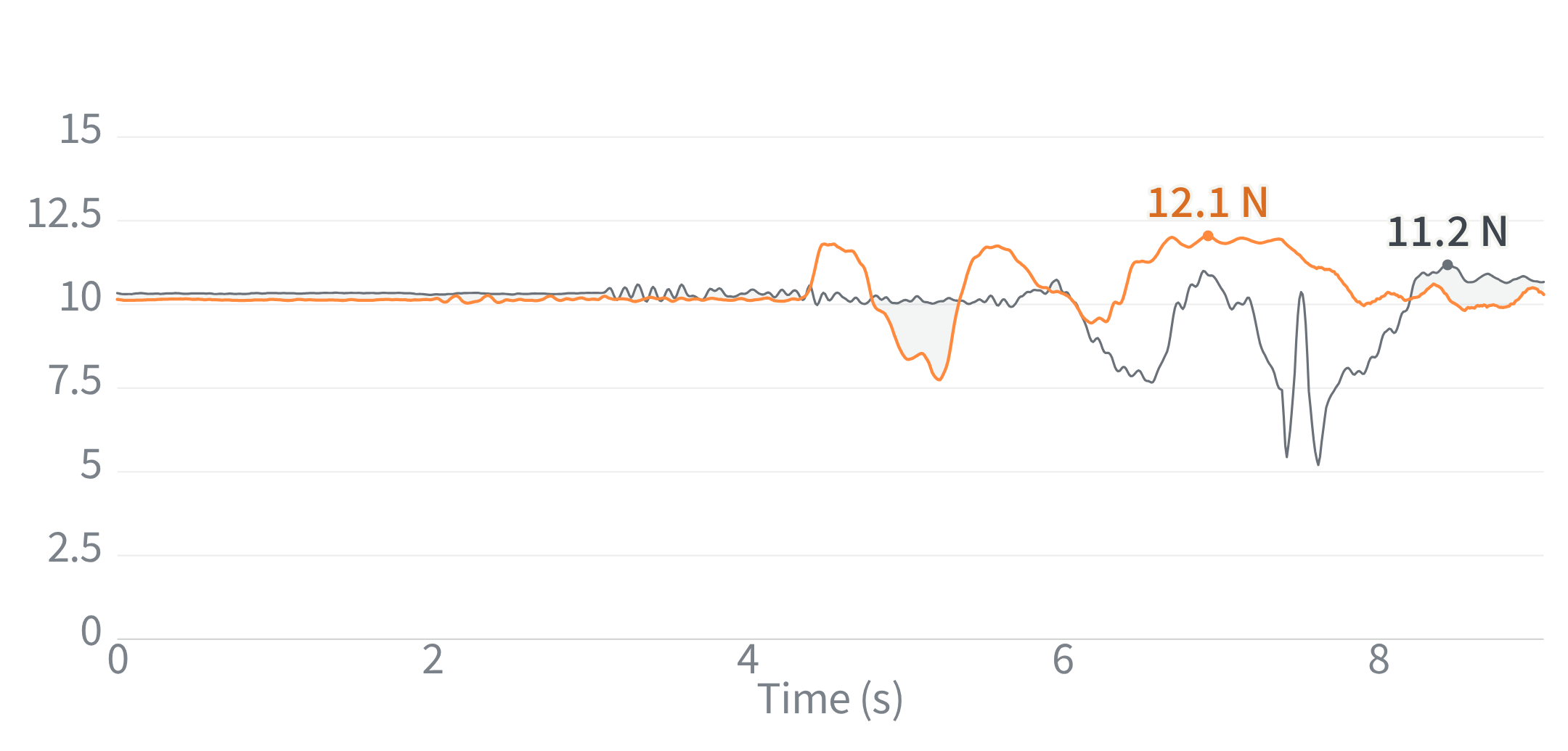}
    \caption{Block stacking (pick and place)}
  \end{subfigure}\hfill
  \begin{subfigure}{0.49\linewidth}
    \includegraphics[width=\linewidth]{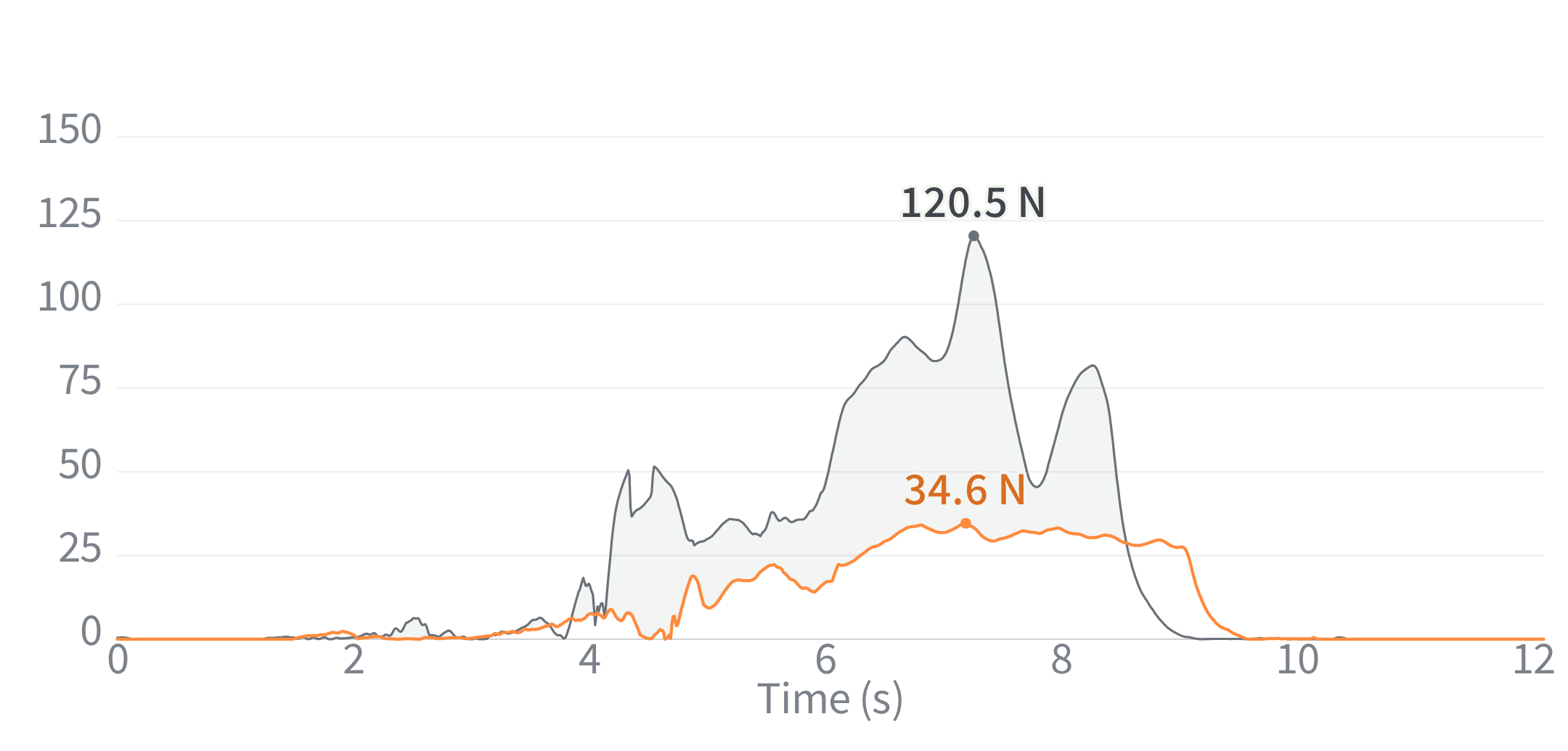}
    \caption{Door opening (hinged object)}
  \end{subfigure}\\[4pt]
  \begin{subfigure}{0.49\linewidth}
    \includegraphics[width=\linewidth]{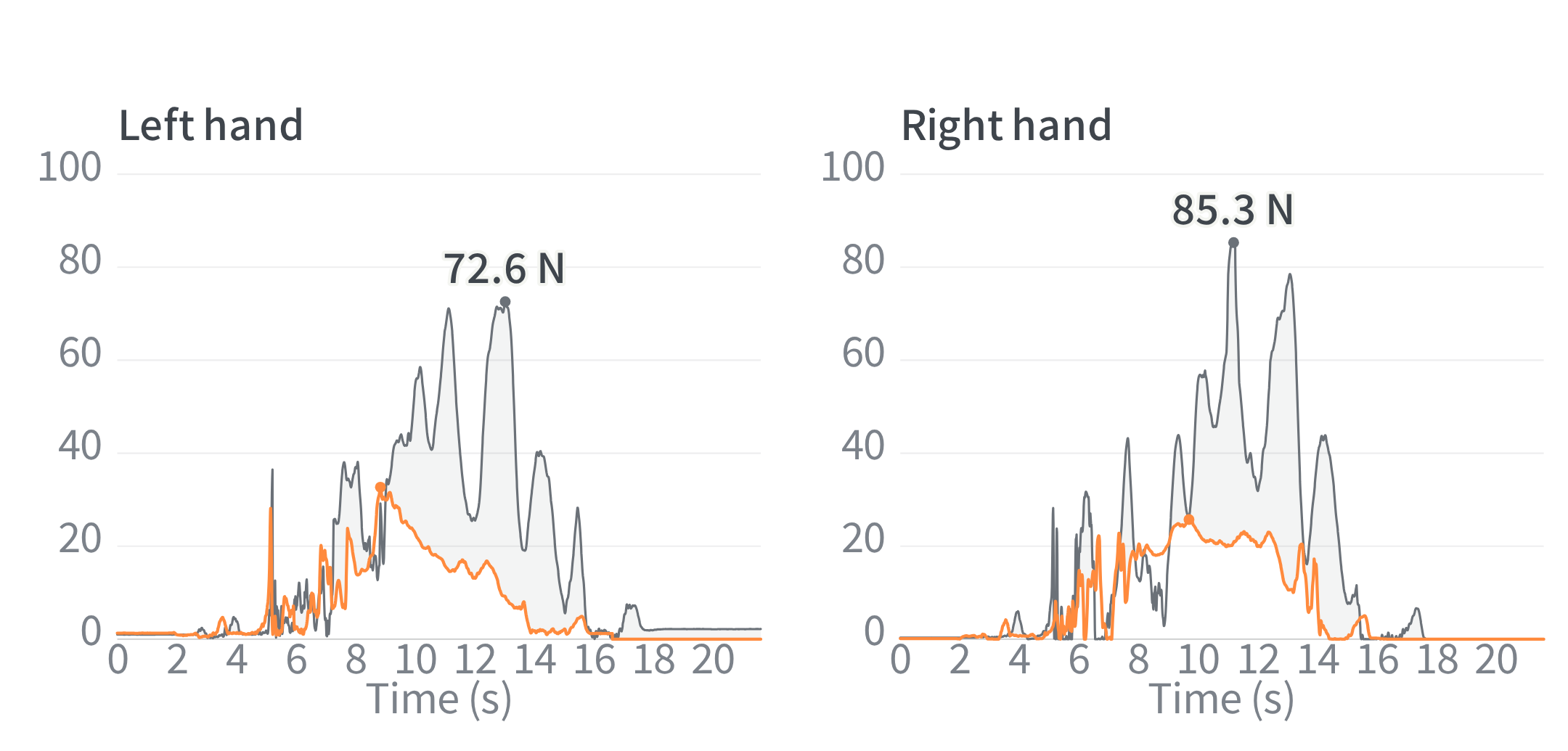}
    \caption{Box carrying (two-hand closed chain)}
  \end{subfigure}\hfill
  \begin{subfigure}{0.49\linewidth}
    \includegraphics[width=\linewidth]{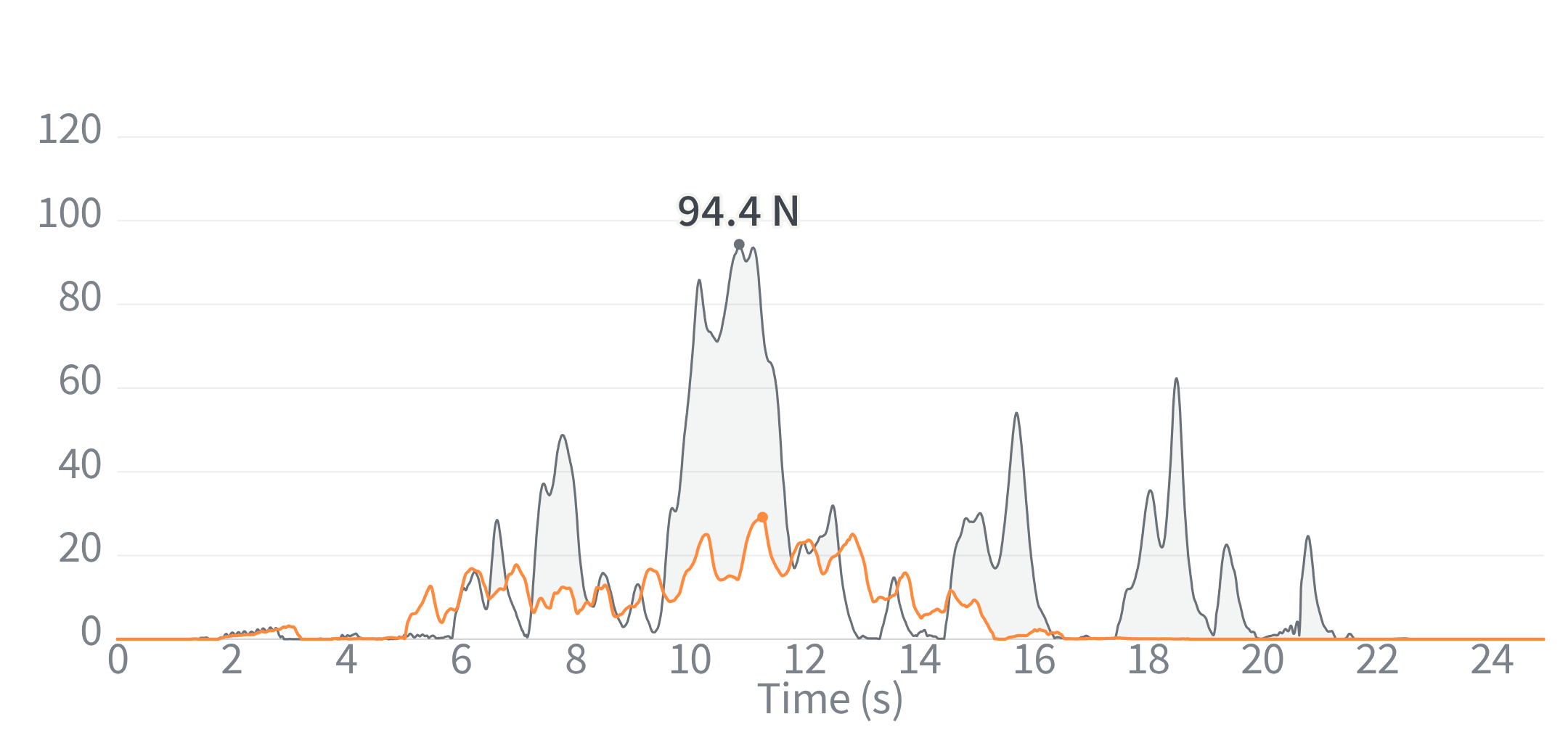}
    \caption{Whiteboard wiping (contact-rich)}
  \end{subfigure}
  \caption{\textbf{Representative end-effector contact-force profiles} under position control (gray) and force control (orange). Vertical axes: contact force magnitude (N).}
  \label{fig:force_curves}
\end{figure}

\begin{table}[!htb]\centering
  \centering
  \caption{\textbf{Share of collection hours needing compliant control}, out of 207{,}030 hours across 20{,}253 task records.}
  \label{tab:compliance_share}
  \begin{tabular}{lrr}
    \toprule
    Category & Hours & Share \\
    \midrule
    Clearly requires compliant control & 38{,}353 & 18.5\% \\
    Benefits from compliant control & 80{,}161 & 38.7\% \\
    \bottomrule
  \end{tabular}
\end{table}
Contact-force data from representative trajectories further illustrate the difference (Figure~\ref{fig:force_curves}). In block stacking, the two modes produce similar force profiles. In door opening, peak contact force drops from 120.5\,N under position control to 34.6\,N under force control, a 71.3\% reduction; in whiteboard wiping it drops from 94.4\,N to 29.2\,N (69.1\%); and in bimanual box carrying, the left- and right-hand peaks fall by about 55.0\% and 69.9\%. In the contact tasks tested, compliant control relieves the conflict between trajectory error and environmental constraints and lowers peak contact force.

The control mode should be chosen according to the task's contact characteristics and positioning requirements. \textbf{For mechanism-constrained, continuous-contact, and bimanual closed-chain tasks, control with appropriate compliance should be preferred to reduce contact-induced corrections and excessive forces; for low-contact tasks that emphasize positional accuracy, stiffer position control may be used.}

\subsection{System Latency: Optimizing the Full Loop and Motion Response}
\label{sec:latency}

\begin{figure}[tbp]
  \centering
  \includegraphics[width=\linewidth]{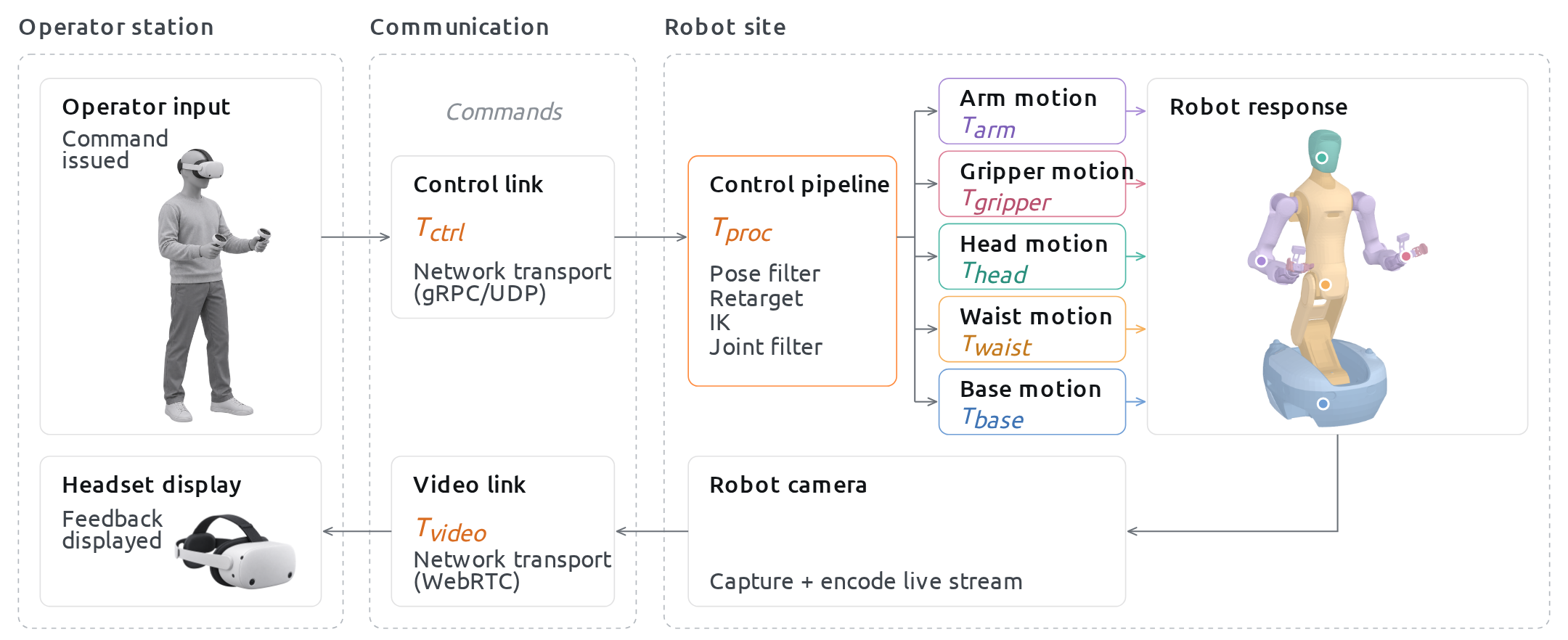}
  \caption{\textbf{Delay accumulation in teleoperation.} The command path runs from operator input through the network link ($T_{\mathrm{ctrl}}$) and control pipeline ($T_{\mathrm{proc}}$: pose filter, retargeting, IK, joint filter) to the body parts, which respond in parallel ($T_i$); visual feedback returns through the robot camera and video link ($T_{\mathrm{video}}$). Timing is measured from command issue to display of the matching feedback.}
  \label{fig:latency_model}
\end{figure}

Teleoperation depends heavily on the closed loop ``operator issues a command, robot executes, visual feedback returns, operator adjusts''~\citep{sheridan1963delay}. System latency arises not only from command transmission and computation, but also from signal filtering, robot dynamic response, and the generation and transmission of visual feedback. Accumulated latency in the control chain leads operators to misjudge the robot's state and make unnecessary overcompensating motions. In a simplified model where the required body parts respond in parallel (Figure~\ref{fig:latency_model}), the total feedback latency is
\begin{equation}
  T_{\mathrm{total}} = T_{\mathrm{ctrl}} + T_{\mathrm{proc}} + \max_{i\in R} T_i + T_{\mathrm{video}},
  \label{eq:latency}
\end{equation}
where $R$ is the set of body parts required by the task, $T_{\mathrm{ctrl}}$ is the command transmission delay, $T_{\mathrm{proc}}$ is the equivalent delay of control processing, $T_i$ is the motion-response delay of part $i$, and $T_{\mathrm{video}}$ is the visual feedback delay. \textbf{The model assumes all parts respond simultaneously, so overall completion is determined by the slowest necessary part.} Robot design should therefore consider whole-body tracking performance: substantially speeding up a single part yields limited overall improvement. \textbf{In our experience, the end effector, waist, and base are the most common efficiency bottlenecks.}

In practice, computation time and motion-response delay differ substantially. We observe that pose filtering, motion retargeting, and joint filtering can each take under 1\,ms per call while the corresponding motion response still lags by tens of milliseconds; some filter configurations take less than 0.01\,ms yet introduce over 200\,ms of phase lag at particular motion frequencies. \textbf{Latency should therefore not be evaluated by algorithm execution time, but measured directly as the time between command input and actual motion response.}

\begin{figure}[tbp]
  \centering
  \includegraphics[width=0.72\linewidth]{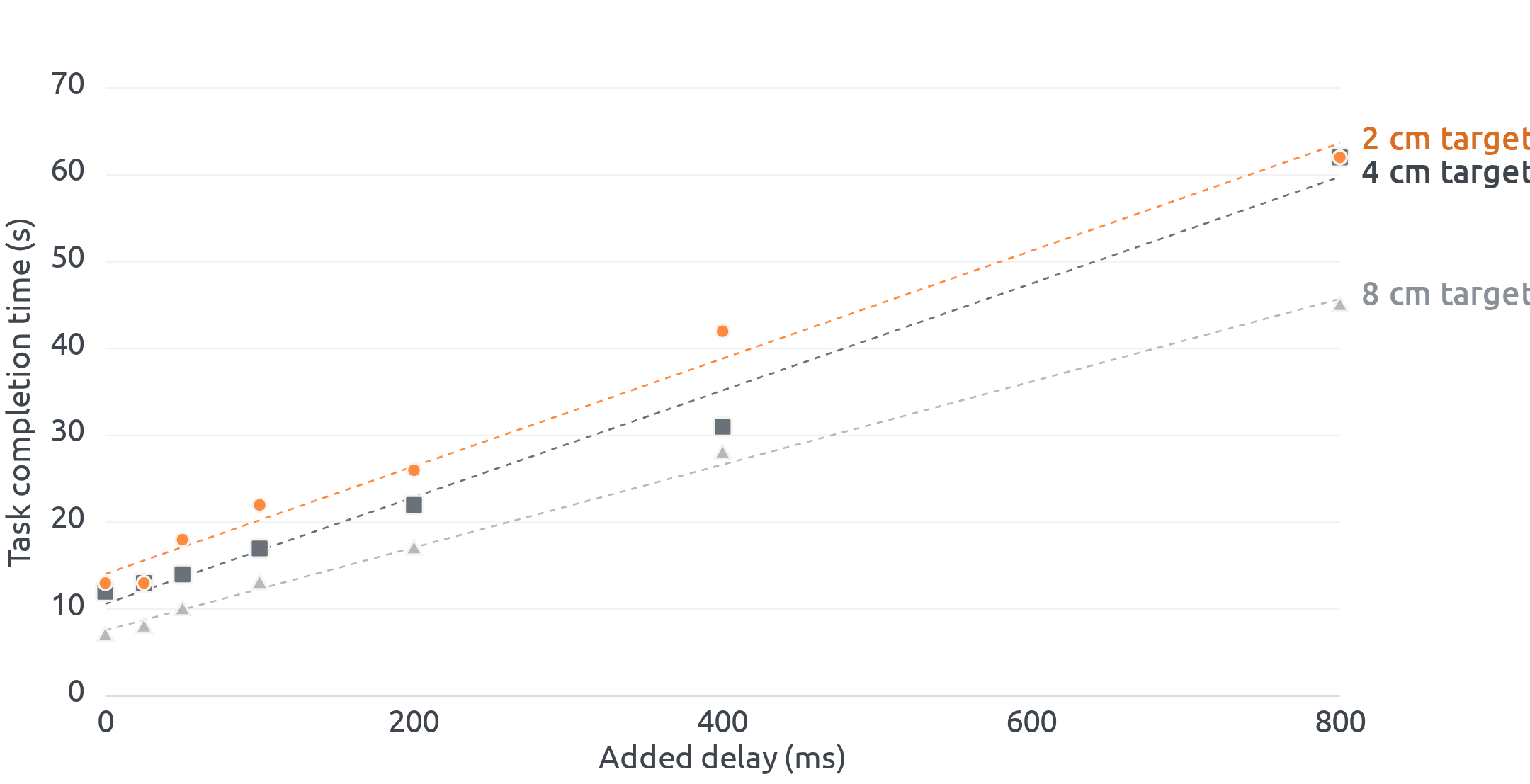}
  \caption{\textbf{Latency sensitivity by target size.} Operators control the robot end effector in simulation to reach randomly generated target positions and orientations as quickly as possible, with target sizes of 2, 4, and 8\,cm and varying injected latency; 10 trials per condition. The horizontal axis is the injected latency; the system runs at 30\,Hz, giving a fixed baseline latency of about 33\,ms.}
  \label{fig:latency_target}
\end{figure}

The effect of latency is especially pronounced in fine tasks. In simulated teleoperation, we set target sizes of 2\,cm, 4\,cm, and 8\,cm, inject varying amounts of extra latency into the control chain, and measure the time operators need to reach the target pose. For the 2\,cm target, increasing extra latency from 0 to 200\,ms raises mean completion time from about 13\,s to 26\,s; at 800\,ms it rises further to about 62\,s (Figure~\ref{fig:latency_target}). Latency makes it hard for the operator to judge the robot's actual position in time, inducing overshoot, waiting for feedback, and reverse correction, all of which lengthen completion time.

The experiment also shows that \textbf{once system latency falls below about 100\,ms, phase lag becomes difficult to perceive and further reductions yield diminishing improvements in completion time}. In simple tasks, end-effector latency affects overall speed almost linearly: each additional 100\,ms of gripper latency adds about 200\,ms to a single pick-and-place, because each pick-and-place contains one opening and one closing, and when the gripper is slower than the arm, the arm must wait for the gripper to settle before moving on. \textbf{The control system should optimize the actual human-in-the-loop response, prioritizing the dominant latency sources---filtering, robot motion response, and visual feedback---rather than merely pursuing lower algorithmic computation time.}

\subsection{Control Peripherals: Balancing Capability and Sustained Workload}
\label{sec:peripherals}

Control peripherals determine how the operator perceives the task environment and conveys intent to the robot. Devices must provide sufficient spatial information and motion-control capability while accounting for wearable burden, fatigue, and efficiency over long collection sessions. Display and motion-input devices should therefore be chosen according to the task and designed together as one interaction system.
\textbf{Display devices shape spatial perception and sustained working capacity.} 2D and 3D screens require no headset and suit long sessions at a fixed station. VR headsets let the viewpoint follow head motion, providing more direct spatial perception and viewpoint control, and suit tasks that require moving, bending, and actively searching for targets; prolonged wear, however, can cause fatigue or motion sickness~\citep{lavioala2000cybersickness}. In our device tests, mean session durations before discomfort were 62.11, 63.14, and 18.71 minutes for 2D screens, 3D screens, and VR headsets, respectively (Table~\ref{tab:dizziness}).

\begin{table}[!htb]\centering
  \centering
  \caption{\textbf{Time to discomfort by display device}: session duration before reported discomfort (min), nine sessions per device. Designated outliers (3D screen: 17 and 25\,min; VR headset: 3 and 8\,min) are excluded from the statistics.}
  \label{tab:dizziness}
  \begin{tabular}{lccc}
    \toprule
    Display & Mean & SD & Sessions used \\
    \midrule
    2D screen & 62.11 & 13.82 & 9 / 9 \\
    3D screen & 63.14 & 9.84 & 7 / 9 \\
    VR headset & 18.71 & 5.35 & 7 / 9 \\
    \bottomrule
  \end{tabular}
\end{table}

\textbf{Motion-input devices must match the task's control DoFs, precision, and range}. Exoskeletons establish a fairly direct mapping between human motion and robot joints or end effectors and, with force feedback, can convey contact information to the operator; their weight, mechanical constraints, and calibration complexity, however, add burden. A SpaceMouse suits small-range six-DoF fine adjustment, but continuous large-range operation typically needs additional control mechanisms. Hand controllers offer a compromise between portability and range: different control modes support fine adjustment, and combining them with headsets, body tracking, and automatic control modules supports more complex whole-body operation.

\textbf{Peripheral selection should jointly consider the task's spatial-perception and motion-control needs and the operator's sustained working capacity.} For fixed stations and long sessions, screens and lightweight input devices are preferable; for tasks that require active viewpoint adjustment and whole-body coordination, VR and body tracking are appropriate; for fine manipulation or tasks that need contact feedback, a SpaceMouse or a force-feedback exoskeleton can be introduced as needed. Peripheral configuration should be driven by task fit and long-term efficiency, not by a pursuit of greater immersion or more input DoFs.

\section{Data Quality: No Missing Information, No Redundancy, Full Consistency}
\label{sec:data}

\textbf{Data quality sets the ceiling of imitation learning; at its core are complete information, motion without redundancy, and end-to-end consistency.}

\emph{Complete information.} If the operator relies on cues that the data do not record, the model cannot learn the causal basis of the action. For instance, if the operator closes the gripper based on a side view while the data record only an occluded front view, the model sees the gripper close ``for no reason.'' Such a gap in the sensing configuration directly undermines learnability; observations must contain all the information needed for the decision.

\emph{No redundant motion.} Ineffective micro-adjustments and exploratory wobbles caused by collection latency are labeled as positive examples. Fitting this behavioral noise wastes model capacity and causes hesitation and oscillation at deployment.

\emph{End-to-end consistency.} Inconsistency stems mainly from two sources. The first is error internal to the data, such as timestamp misalignment that offsets state and action by several frames, or inaccurate hardware calibration, which the model learns as if it were real structure. The second is cross-domain error at inference, such as different joint zero positions between the deployment and collection robots, or changes in inference or motion-control parameters; such heterogeneous errors are out-of-distribution perturbations the model has never seen and readily cause compounding errors and task failure.

\begin{figure}[tbp]
  \centering
  \includegraphics[width=\linewidth]{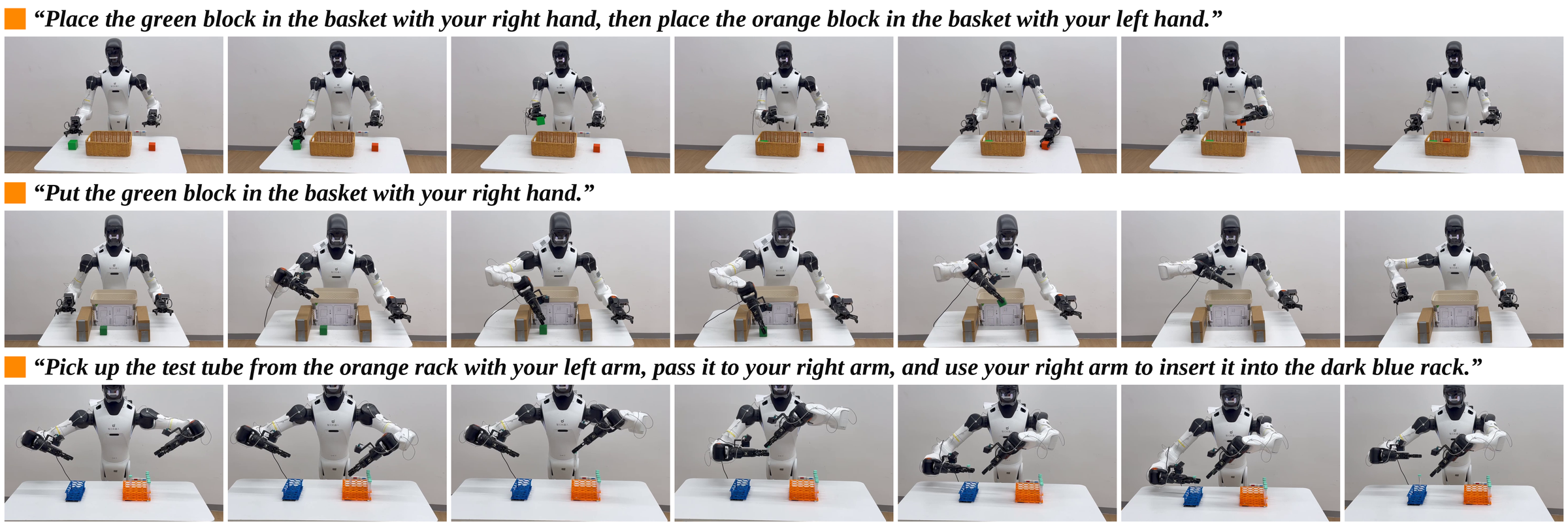}
  \caption{\textbf{Evaluation tasks used in the data-quality experiments:} bimanual block transfer (\S\ref{sec:redundancy}), out-of-view block pick-and-place (\S\ref{sec:missing}), and test-tube handover and insertion (\S\ref{sec:consistency}).}
  \label{fig:tasks}
\end{figure}

In this section we quantify the effect of each of these three issues on downstream post-training. All experiments use our own collected data and evaluate policies obtained by post-training \pizerofive~\citep{black2025pi05}. Figure~\ref{fig:tasks} shows the evaluation tasks.

\subsection{Missing Information: Visual State Must Explain the Action}
\label{sec:missing}

A demonstration records what the robot did, but not necessarily why the operator did it. When an action depends on a visual cue, the model must have access to that information before the action occurs. The absence of such a signal cannot be compensated by more data.

\begin{figure}[tbp]
  \centering
  \includegraphics[width=\linewidth]{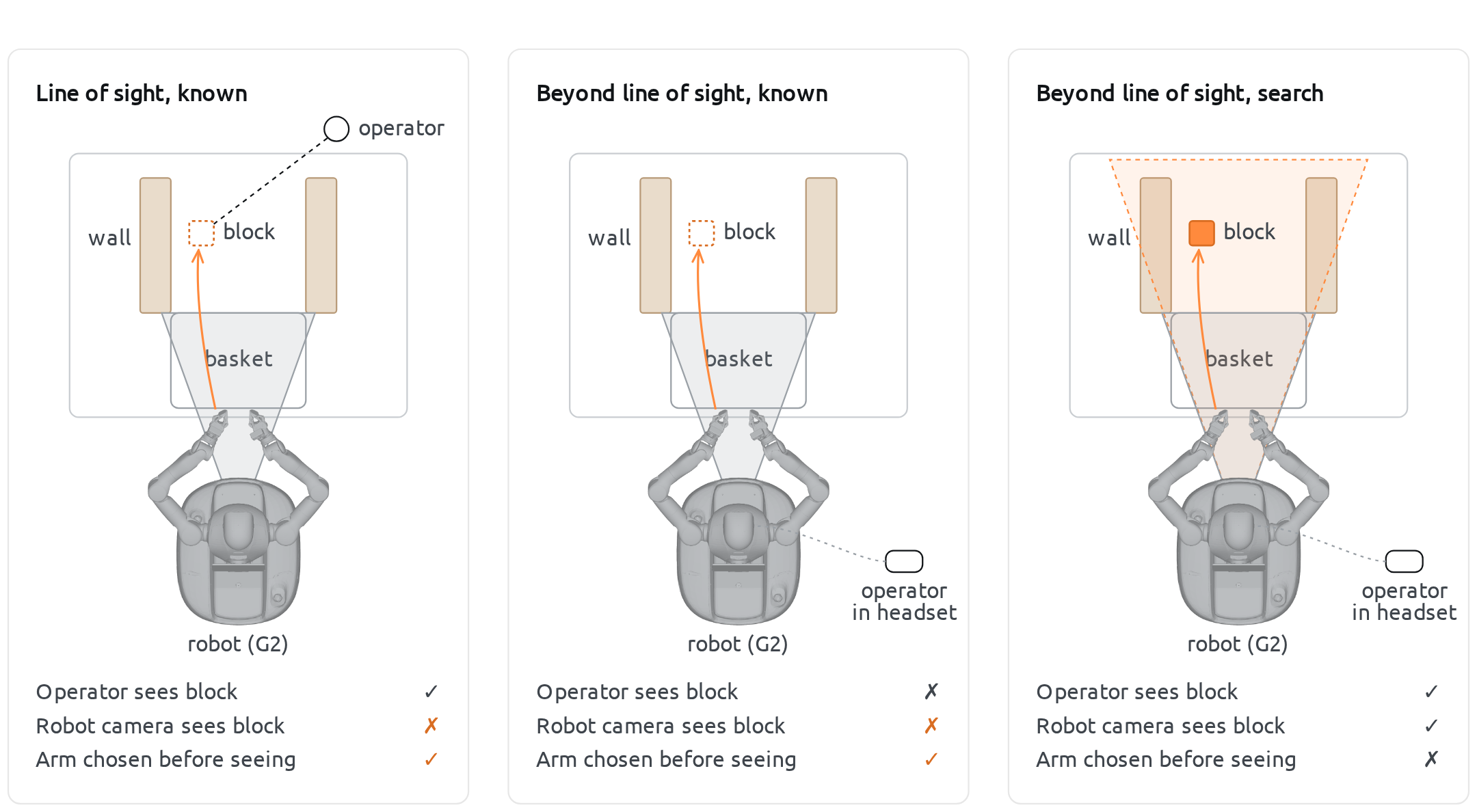}
  \caption{\textbf{Demonstration setups.} LOS-known: the operator sees the block, the robot camera does not, and the arm is chosen before seeing. BLOS-known: neither operator nor robot camera sees the block initially, and the arm is still chosen before seeing. BLOS-search: the robot camera sees the block before the arm is chosen.}
  \label{fig:info_setups}
\end{figure}

To test this, we design a simple task: a block is placed outside the robot's initial field of view, and the robot must pick and place it with the nearer hand. We compare three demonstration protocols (Figure~\ref{fig:info_setups}). In \emph{LOS-known}, the operator stands in the scene and can see the block directly, and so reaches with the nearer hand without looking; the robot's initial observation lacks this cue, leaving the action without a causal basis. In \emph{BLOS-known}, the operator's view is aligned with the robot's, but the operator still reaches directly based on prior knowledge of the block position. In \emph{BLOS-search}, the operator first has the robot lean over to look, confirms the block's location, and only then chooses which hand to use.

The results differ sharply (Table~\ref{tab:info_results}). In arm-selection accuracy, BLOS-known (53\%) and LOS-known (about 50\%), both lacking the initial visual cue, are close to random guessing, whereas BLOS-search, which includes active search, reaches 100\%. Its 47-point improvement clearly exceeds trial-to-trial variation (30 trials per condition; 95\% confidence interval of the difference 27--64 points). In subsequent pick-and-place success, BLOS-search again performs best, rising from 40\% for BLOS-known to 60\%, while LOS-known achieves only 33\%.
\textbf{For tasks with occlusion or blind spots, demonstrations must capture the complete ``observe, decide, act'' sequence, so that information-gathering motions such as leaning, turning the head, and adjusting the viewpoint enter the training data, and the image history needed for the decision is retained.} This is not a mere performance optimization but decides task success: when key decision information is absent, the model can only guess from scene priors, whereas once demonstrations include active observation, an otherwise ambiguous action choice acquires a clear visual basis.

\begin{table}[!htb]\centering
  \centering
  \caption{\textbf{Arm choice and pick-and-place success} for the three demonstration protocols, 30 trials each. A random arm choice is correct 50\% of the time.}
  \label{tab:info_results}
  \begin{tabular}{lcc}
    \toprule
    Protocol & Correct arm (\%) & Pick-and-place success (\%) \\
    \midrule
    LOS-known & 50 & 33 \\
    BLOS-known & 53 & 40 \\
    BLOS-search & 100 & 60 \\
    \bottomrule
  \end{tabular}
\end{table}

In imitation learning this problem is known as \emph{expert--learner observation mismatch}~\citep{swamy2022unobserved,dehaan2019causal}: the operator uses extra information to make the right decision, while the model lacks the corresponding observation. LOS collection is particularly prone to it, since operators acquire environmental information beyond the robot's cameras with their own eyes and then embed this knowledge implicitly in their actions.

BLOS collection provides a direct check: by restricting the operator's decision basis to the visual information the robot actually records, it becomes easier to discover which decisions lack observational support. The BLOS-known condition also shows that aligning viewpoints is not enough; active search is still needed for decision cues to actually enter the data.

\textbf{Therefore, tasks involving occlusion, out-of-view targets, and branching decisions should preferably be collected BLOS, with active observation written into the demonstration protocol.} Given the burden of prolonged VR wear, remote screen-based teleoperation can also be used, letting operators complete the task through the same camera views the model receives.

\subsection{Inconsistency: Calibration and Collection--Inference Alignment}
\label{sec:consistency}

The model takes vision and proprioceptive state as input and predicts the next action. If these are systematically misaligned in the data, the model may learn the misalignment as if it were a regularity. Timestamp offsets make state and action fail to correspond; joint zero-offset miscalibration pairs the same image with different proprioceptive states. Both can degrade model performance.

\begin{figure}[tbp]
  \centering
  \includegraphics[width=\linewidth]{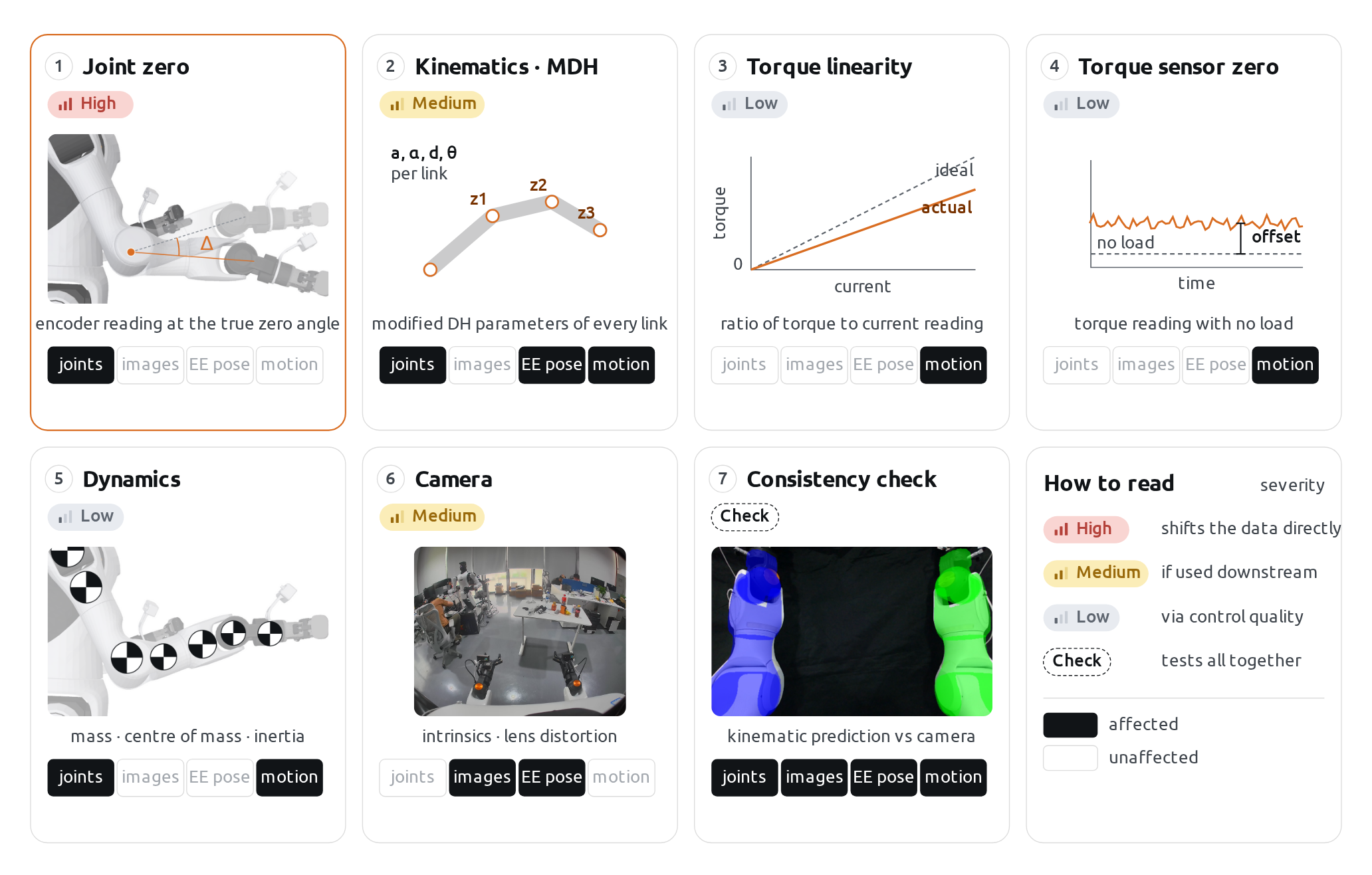}
  \caption{\textbf{Calibration items affecting robot data}: joint zero, kinematics (MDH), torque linearity, torque-sensor zero, dynamics, and camera, followed by a whole-robot consistency check. Joint zero (orange outline) is the item varied in our experiment.}
  \label{fig:calib_types}
\end{figure}

Categorizing our robot data, we find the relevant factors to be joint zero positions, kinematic parameters, torque sensors, dynamics, and camera intrinsics and extrinsics, with overall machine quality finally judged by a whole-robot consistency check (Figure~\ref{fig:calib_types}). Their effects differ. For VLA and world-action-model (WAM) training data, the joint zero position most directly affects the state representation, while the other calibrations act mainly through pose computation, control tracking, or downstream data processing. We therefore focus on joint zero positions and analyze how zero offsets affect model inference.

Prior work has shown that predicting delta actions outperforms predicting absolute actions~\citep{deltaaction2026}, but it has not examined whether, under a delta-action representation, joint zero offsets degrade model performance (see Appendix~\ref{app:delta}). Can data with different zero calibrations be mixed in training? If so, how much inconsistency can the model tolerate?

\begin{figure}[tbp]
  \centering
  \includegraphics[width=\linewidth]{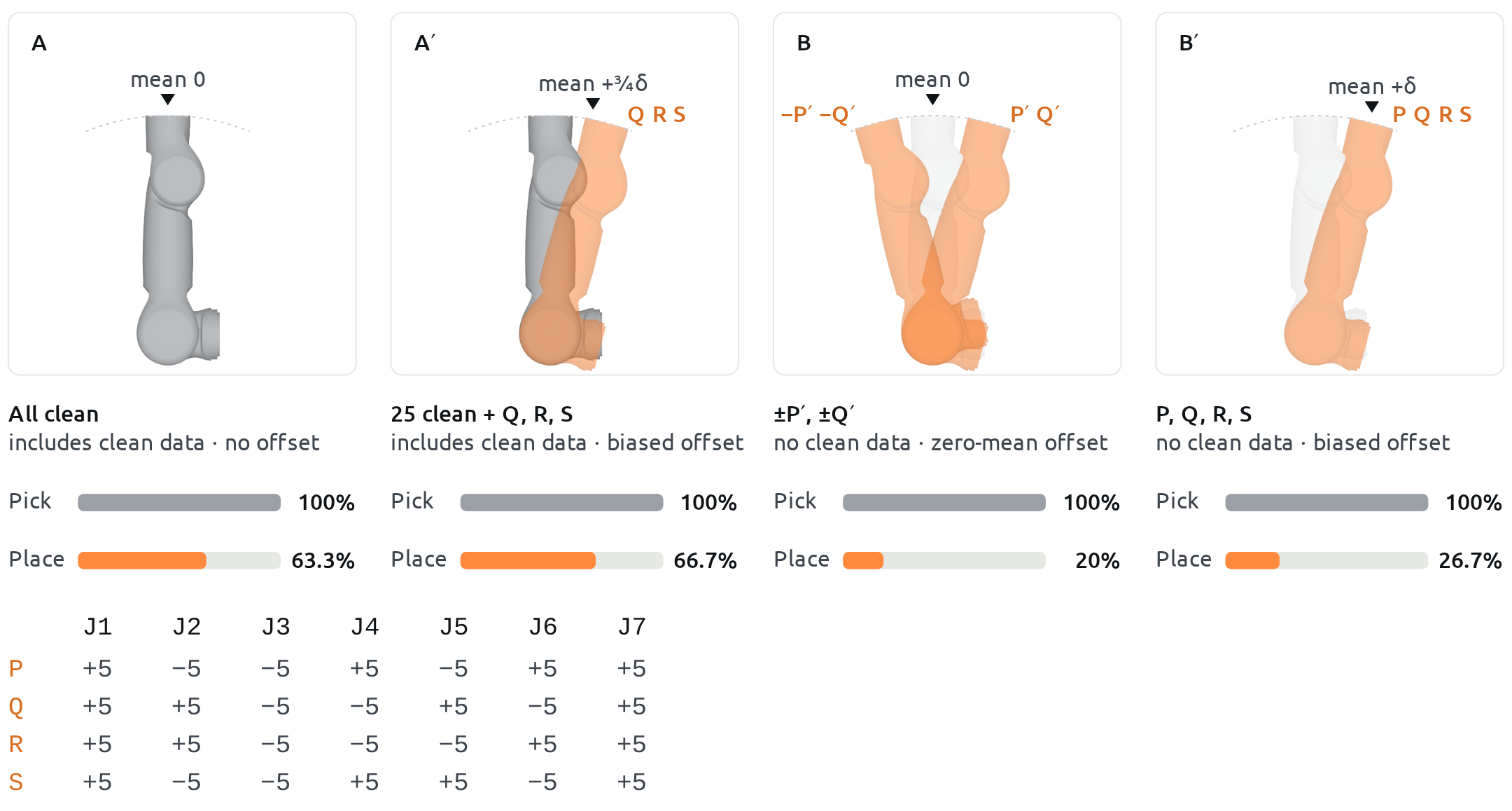}
  \caption{\textbf{Training mixtures for the zero-offset experiment.} Each mix has 100 episodes in four groups of 25. A: all clean. \Aprime: 25 clean plus groups Q, R, S with biased offsets. B: $\pm P$, $\pm Q'$, no clean data, zero-mean offsets. \Bprime: P, Q, R, S, no clean data, biased offsets. The table lists per-joint offsets (degrees) on J1--J7; bars show real-robot pick-and-place success.}
  \label{fig:calib_mixes}
\end{figure}

To emulate different zero-offset scenarios, we start from complete demonstrations and inject fixed offsets batch-wise into the 14 joint states. Two variables---whether clean data is included and whether offsets cancel out---define four training mixtures (Figure~\ref{fig:calib_mixes}): \textbf{A}, all clean data; \textbf{\Aprime}, 25 clean episodes plus 75 episodes with same-sign (biased) offsets; \textbf{B}, no clean data, with zero-mean offsets of opposite signs; and \textbf{\Bprime}, no clean data, with biased offsets. The test task has the left arm take a test tube from an orange rack and hand it to the right arm, which inserts it into a dark-blue rack. The early grasping phase is relatively coarse, while the later handover and insertion require more precise bimanual coordination.

\Aprime, which keeps only 25\% clean data, reaches 66.7\% success in the precise phase, essentially on par with the all-clean A at 63.3\%. With 30 trials per group, the 95\% confidence half-width of each success rate is about 14--16 points; in this experiment, a large share of zero-offset data did not damage performance.

The real gap is between \Aprime and \Bprime. The two share 75\% of their data and differ only in replacing the remaining 25\% clean data with another offset group, yet success drops from 66.7\% to 26.7\%, a 40-point difference (95\% CI of the difference about 15--59 points) that exceeds trial variation. This suggests that a small amount of data aligned with the inference zero position may serve as a crucial reference. Configuration B provides another counterexample: even though positive and negative offsets cancel out overall, success is only 20\%. A zero-mean offset does not guarantee a consistent state--action relationship within each episode. \textbf{Hence, historical data need not be discarded wholesale because of differing zero-calibration batches, but the dataset should contain baseline trajectories aligned with the inference robot.}

\begin{figure}[tbp]
  \centering
  \includegraphics[width=\linewidth]{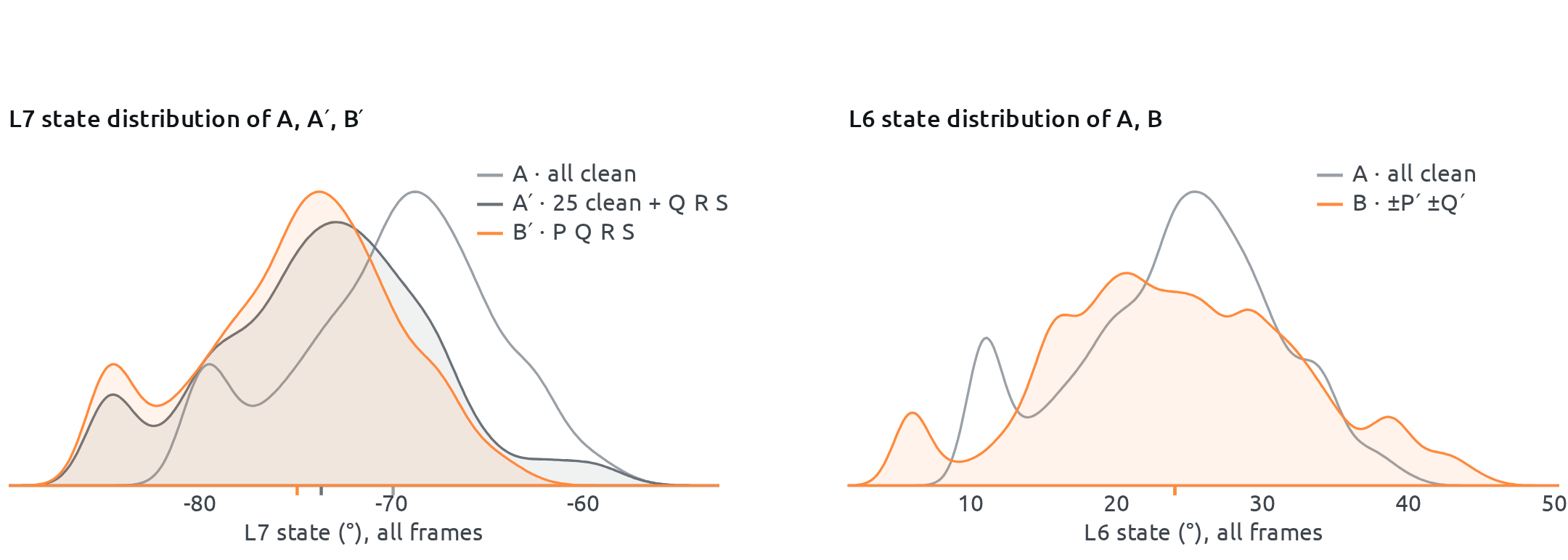}
  \caption{\textbf{How each mixture shifts the joint-state distribution}, shown for two representative joints across the four training mixes.}
  \label{fig:calib_state}
\end{figure}

Why did part of the offset data not hurt performance? We attribute this mainly to a change in how the model weighs vision against proprioception (Figure~\ref{fig:calib_state}): it focuses more on visual feedback than on proprioceptive values. In heterogeneous data with joint-zero offsets, the same joint reading may correspond to entirely different physical poses and actions across batches, introducing ambiguity in proprioceptive space. If vision suffices to close the error loop between end effector and target, this state perturbation can weaken overfitting to absolute joint values, acting like data augmentation; conversely, if the task depends heavily on precise fusion of vision and proprioception, the ambiguity can severely disrupt decisions. The behavior of \Aprime is consistent with the former interpretation. In contrast, in our test with a joint-zero calibration offset between the collection and inference robots, all 30 trials failed. We conclude that while training data may come from different robots, zero-offset error on the inference robot severely damages performance.

\textbf{Collection--inference zero alignment should therefore be a mandatory check before model deployment}: record calibration versions, verify joint readings at a standard pose, and compensate each joint's zero offset individually. \textbf{For existing mixed data, preferentially retain reference samples consistent with deployment.} The goal of calibration is not to make all historical data look identical, but to ensure that the state--action relationship the model learns still holds at inference time.

\subsection{Redundancy: Meaningless Human-Induced Motion in Actions}
\label{sec:redundancy}

The embodiment limits, control mappings, and latency discussed above often force operators to pause, detour, or correct repeatedly. These compensatory motions help complete the current task, yet they are written into the training data as positive examples. How much of this redundant motion does the model learn, and in what form does it appear? \textbf{Our experiments suggest that motion trajectories may expose data contamination earlier than final task success does.}

\begin{figure}[tbp]
  \centering
  \includegraphics[width=\linewidth]{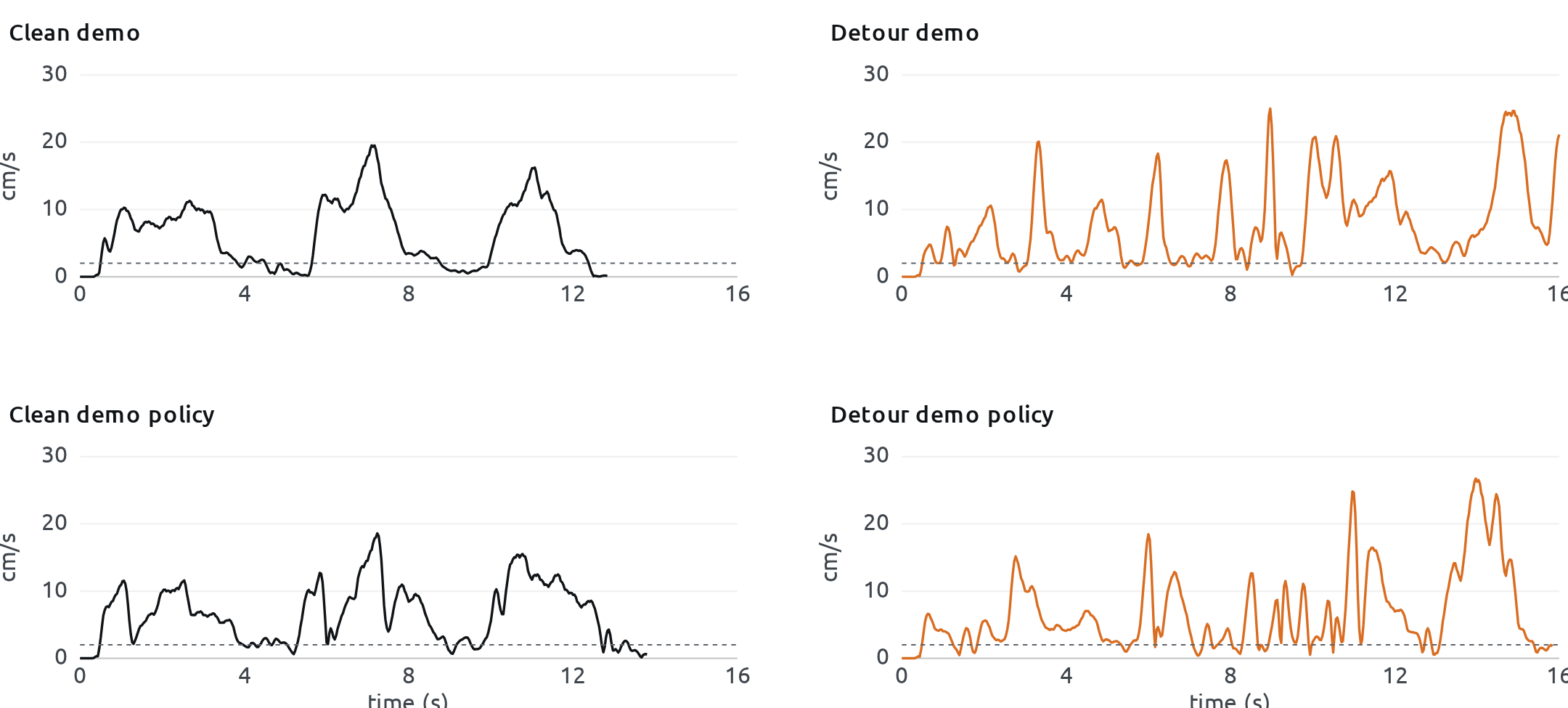}
  \caption{\textbf{Speed profiles of clean and detour demonstrations} (top) and of the corresponding policies at inference (bottom).}
  \label{fig:redundancy_profiles}
\end{figure}

We use a bimanual block-transfer task: the right arm picks and places a block, then the left arm repeats the operation. Four models are each trained on 100 demonstrations, of which 0\%, 20\%, 50\%, or 80\% are replaced by trajectories containing pauses and detours. Compared with clean demonstrations, these trajectories have on average 33\% longer wrist paths, about 1.8$\times$ the joint jerk, and 1.6 rather than 1.1 single-arm pauses longer than 0.3\,s (Figure~\ref{fig:redundancy_profiles}).

Real-robot inference reveals an interesting asymmetry: \textbf{the model smooths some of the motion roughness but retains most of the redundant path} (Table~\ref{tab:detour_metrics}). From 0\% to 80\% mixture, the demonstrations' mean wrist path grows by 0.215\,m and the policy's by 0.194\,m, i.e., the policy retains about 90\% of the added distance; by contrast, only about 45\% of the increase in joint jerk is retained. The model partially smooths abrupt accelerations but still tends to follow the detours in the demonstrations. Data cleaning should therefore target redundant corrections in the trajectory itself; improving smoothness alone still leaves much ineffective motion.
Pauses and corrections can also take different forms at inference. Low-speed events longer than 0.3\,s do not vary consistently with the mixture ratio, but speed peaks per meter of path rise from 10.8 to 19.1 and direction reversals from 3.2 to 10.0. Even at 20\% mixture, where mean joint jerk is still close to the clean model, the velocity mismatch at boundaries between consecutive action chunks already rises from 0.36 to 0.58, reaching 0.78 at 80\% (Table~\ref{tab:detour_metrics}). Some motion defects first appear as repeated speed changes and planning discontinuities, and only later as growth in overall jerk.

\begin{table}[!htb]\centering
  \centering
  \caption{\textbf{Motion metrics versus detour share in training.} Demo: the collected demonstrations of each mixture; Policy: real-robot rollouts of the policy trained on it. Values are means across runs. Handover mismatch is $|v_{\mathrm{new}}-v_{\mathrm{old}}|/|v_{\mathrm{old}}|$ at the boundary between consecutive action chunks.}
  \label{tab:detour_metrics}
  \begin{tabular}{llcccc}
    \toprule
    & & \multicolumn{4}{c}{Detour share in training} \\
    \cmidrule(lr){3-6}
    Metric & Source & 0\% & 20\% & 50\% & 80\% \\
    \midrule
    Joint acceleration RMS (rad/s$^2$) & Demo & 0.26 & 0.31 & 0.40 & 0.48 \\
     & Policy & 0.25 & 0.24 & 0.30 & 0.32 \\
    Joint jerk RMS (rad/s$^3$) & Demo & 3.29 & 3.82 & 4.73 & 5.46 \\
     & Policy & 3.23 & 3.28 & 3.98 & 4.21 \\
    Wrist path length (m) & Demo & 0.83 & 0.88 & 0.97 & 1.04 \\
     & Policy & 0.94 & 0.97 & 1.06 & 1.13 \\
    Brief slowdowns (per 10 episodes) & Demo & 23.0 & 22.5 & 25.3 & 28.3 \\
     & Policy & 24.0 & 23.3 & 37.0 & 32.0 \\
    \midrule
    Handover mismatch & Policy & 0.36 & 0.58 & 0.75 & 0.78 \\
    Time to final release (s) & Policy & 25.3 & 28.6 & 29.5 & 32.6 \\
    \bottomrule
  \end{tabular}
\end{table}

One explanation worth further verification lies in the policy's replanning mechanism~\citep{zhao2023aloha,black2024pi0}: the model periodically generates new action chunks, and after being influenced by pause and correction trajectories, consecutive chunks may differ more in direction and speed. We conjecture that in the training set, pauses and disturbances make the same state co-occur with more uncertain actions, amplifying velocity differences between adjacent outputs.

These motion differences ultimately translate into execution efficiency (Table~\ref{tab:detour_metrics}). As the share of detour data increases from 0\% to 80\%, the time from motion onset to final release grows from 25.3\,s to 32.6\,s, about 29\% longer. Since the mean wrist speed during motion changes little, the extra time mainly accompanies longer paths and more waiting. Even when the task succeeds, the policy may pay a substantially higher time cost.

Acceptance of demonstration data should therefore consider \textbf{path efficiency, motion continuity, and task duration} together. Necessary motions arising from task needs such as observing a target or avoiding obstacles should be kept along with their decision basis; repeated corrections caused by joint limits, awkward mappings, or control lag should be addressed by improving the collection system first, and kept out of the training set as far as possible.

An extra motion made to compensate for a system defect may, after training, become a behavior the robot repeats. \textbf{The operating cost in the collection system may ultimately turn into a cost the robot keeps paying during autonomous execution.}

\section{Related Work}
\label{sec:related}

\paragraph{Data sources for manipulation learning.}
Learning from demonstration~\citep{argall2009survey} underpins diffusion-based and generalist VLA policies~\citep{chi2023diffusion,ze2025generalizable,black2024pi0,black2025pi05}. Their data supply spans real-robot corpora~\citep{oxe2024,khazatsky2024droid,agibotworld2025,wang2026remora}, egocentric datasets~\citep{ego4d,egodex,egoverse} and human-to-robot learning~\citep{egomimic,ego_cevr,egovla,humanoid_human_policy}, portable UMI-style collection~\citep{umi,fastumi,dexumi,hil_umi}, and simulation-based demonstration generation~\citep{mimicgen,robotwin}. Together, these directions broaden scale and diversity, while leaving a common requirement: observations and actions must provide consistent, learnable supervision.

\paragraph{Teleoperation and collection methods.}
Low-cost bimanual and whole-body systems~\citep{zhao2023aloha,fu2024mobilealoha}, immersive visual and haptic interfaces~\citep{cheng2024opentelevision,bunny_visionpro}, and coordinated mobile manipulation systems~\citep{mobile_television,panovla} expand collection capabilities. Motion reconstruction and retargeting connect human movement to robot execution~\citep{megovista,gmr}. Autonomous acquisition and trajectory generation~\citep{roboclaw,fieldgen}, deliberate perturbations~\citep{adc}, and human corrective guidance~\citep{gcent} further improve collection efficiency or informativeness. We complement these approaches by measuring how interface parallelism, mapping, latency, and peripheral burden affect operator throughput and demonstration continuity.

\paragraph{Embodiment, calibration, and action representations.}
Embodiment co-design~\citep{transformer_transformer} and robot-world and hand-eye calibration~\citep{handeye_calibration} address physical capability and coordinate alignment, while kinematics-aware policies and action-space studies examine joint--Cartesian and spatial--temporal representations~\citep{implicit_kinematic_policies,deltaaction2026}. Our analysis connects orientation-aware workspace and joint margin to collection demands, and examines delta joint actions under heterogeneous calibration across collection and deployment robots.

\paragraph{Demonstration quality and system effects.}
Prior work relates imitation-learning performance to demonstration quality, operator proficiency, action divergence, and state diversity~\citep{mandlekar2021matters,belkhale2023dataquality}. Observation mismatch and causal confusion~\citep{swamy2022unobserved,dehaan2019causal} explain why successful demonstrations can remain difficult to learn from; interactive correction~\citep{ross2011dagger} offers one remedy. Established studies of impedance control~\citep{hogan1985impedance}, teleoperation delay~\citep{sheridan1963delay}, and cybersickness~\citep{lavioala2000cybersickness} provide foundations for analyzing collection-system constraints. We connect these concerns by tracing how embodiment and control choices increase operator effort and introduce missing information, calibration inconsistency, or redundant motion into policy training.

\section{Conclusion}
\label{sec:conclusion}

From embodiment design through the control system to data quality, our analysis points to a single chain: \textbf{the design of the collection system determines how much extra effort operators must spend, and that effort can in turn shape the training data and the final model behavior.}

At the embodiment level, the effective workspace should be evaluated in terms of the positions, end-effector orientations, and joint margins a task requires, with waist DoFs and end-effector length configured accordingly. At the control level, body parts should be operated in parallel, compliant control should be chosen for contact tasks, and the full human-in-the-loop path, including visual feedback, should be optimized to reduce waiting, switching, and repeated correction. At the data level, key observations must explain the actions, collection and inference must share a consistent state reference, and ineffective compensatory motions in demonstrations must be controlled.

These findings suggest a practical engineering path: \textbf{before scaling collection, validate the embodiment's effective workspace, the operator's efficiency, and the data's learnability on representative tasks.} Beyond task success, record completion time, repeated corrections, trajectory length, and motion continuity, and check data quality through small-scale model training and real-robot testing. Fix the system bottlenecks these reveal first, then scale up collection.

Scaling collection requires continually raising throughput, but also reducing the wasted operations and downstream rework behind each demonstration. Improving the collection system promises to raise both data-production efficiency and the quality of the motions models learn.

\paragraph{Limitations and future work.}
Our experiments have limited scope. The embodiment analysis relies mainly on URDF models, discretized IK, and representative morphologies; real workspaces are further constrained by collisions, dynamics, and hardware limits. The control-system experiments cover a few representative tasks and a limited set of operators and device configurations; other robot platforms and proficiency levels may yield different results. The data-quality experiments focus on specific tasks, a single policy model, and one action representation, and some use artificially injected offsets or a single training run. They reveal concrete data failure modes, but the magnitude and generality of these effects deserve validation on more tasks and policy architectures. Future work will further connect collection efficiency, trajectory quality, and model performance: under the same tasks and collection budget, we will compare data produced by different system configurations and the policies trained on them, forming a complete evaluation pipeline spanning collection, quality inspection, training, and real-robot acceptance.

\clearpage
\phantomsection
\addcontentsline{toc}{section}{Contributions}
\section*{Contributions}
\label{sec:contributions}
\noindent\textbf{Core Contributors:} Che Tsang Calvin$^{*}$, Xingyu Chen$^{*}$, Bowen Jiang$^{*}$, Naixiang Gao$^{*}$, Xiongfeng Cai

\smallskip\noindent\textbf{Team Lead:} Wenhao Wang$^{*\dagger}$, Li Liu

\smallskip\noindent\textbf{Contributors:} Enlin Gu, Ruiying Li, Peize Li, Mianrui Zhu, Haomiao Zhang, Xin He, Minhui Yu, Jun Zhao, Zihao Zhu, Yang Chen, Jingmei Jin, Keru Zhou, Chen Wang, Chang Zhou, Zhiwen Hou, Di Yang, Wenjun Sun, Fan Yang, Liyan Zhang, Yi Lu, Haojun Ma, Jun Zhu, Pengxin Zha, Zhaosheng Wang, Xin Zhao, Xindong He, Chiming Liu, Jianheng Song, Jiayao Ma

\smallskip\noindent\textbf{Sponsor:} Maoqing Yao, Guanghui Ren, Qian Wang, Wei Shan, Yang Yang

\medskip\noindent{\small $^{*}$\,Equal Contribution. \quad $^{\dagger}$\,Corresponding Author.}

{
\small
\bibliography{ref}
}

\clearpage
\appendix
\titleformat{\section}
  {\sectionfont\sffamily\bfseries\color{reportorange}}
  {Appendix~\thesection}{0.8em}{}
\section{Why We Use Delta Joint Actions}
\label{app:delta}

\citet{deltaaction2026} provide an empirical analysis of action-space design, highlighting the benefits of delta actions and joint-space representations for modern policy backbones, including diffusion-based models. Motivated by these findings, we examine delta joint actions in data collected from multiple robots with different joint zero calibrations. The following analysis illustrates how this representation cancels constant zero offsets in action labels, while leaving their effects on the observed state.

\subsection{Where Do Data from Different Robots Disagree?}

\begin{figure}[tbp]
  \centering
  \includegraphics[width=\linewidth]{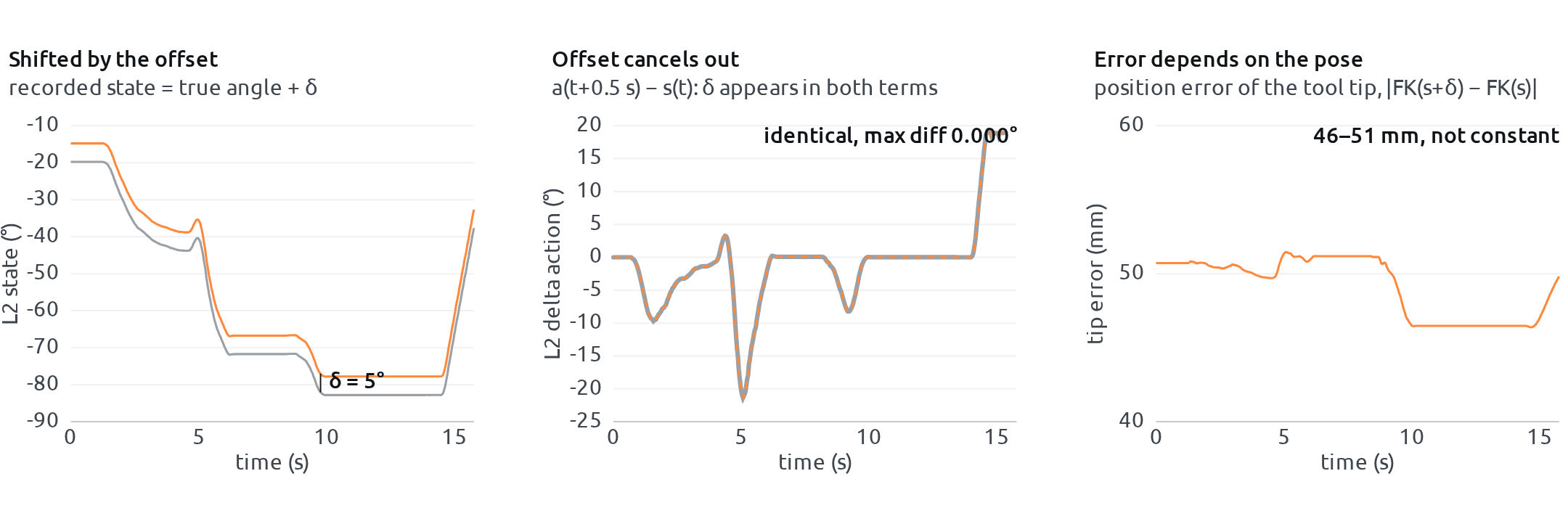}
  \caption{\textbf{What a zero-position offset does to the data.} The same recorded episode, shown clean and with a fixed offset on every joint. The offset shifts the recorded state, cancels in the delta action, and the resulting end-effector error depends on the pose. Gray: clean recording; orange: the same recording with the offset.}
  \label{fig:app_offset}
\end{figure}

Suppose we have four mechanically identical robots with the same kinematic model, camera configuration, and actuator performance, differing only in joint zero calibration. We group them into three calibration levels: A is accurately calibrated, B has a slight offset, and C a noticeable one. We describe robot $i$ by
\begin{equation}
  q_i^{\mathrm{obs}} = q^{\mathrm{true}} + b_i .
\end{equation}
If all robots are asked to execute exactly the same motion, e.g., move the gripper to a given position on the table, they physically reach the same place, but their recorded states differ (Figure~\ref{fig:app_offset}). A typical policy learns
\begin{equation}
  \pi_\theta(I_t, s_t) \rightarrow a_t ,
\end{equation}
where $I$ is the image, $s$ the robot state, and $a$ the action label.

\textbf{If the action is an absolute joint target, the zero offset affects both state and action.} With position control through encoders carrying the same zero offset, a true target angle of 30$^\circ$ may be recorded as 30$^\circ$ on A but must be recorded as 34$^\circ$ on C; both the action and the state are biased.

\textbf{If the action is a joint increment, the offset cancels in the action and affects only the state.} Here
\begin{equation}
  \Delta q_i = q_{i,t+1}^{\mathrm{obs}} - q_{i,t}^{\mathrm{obs}}
  = \left(q_{t+1}^{\mathrm{true}} + b_i\right) - \left(q_t^{\mathrm{true}} + b_i\right)
  = q_{t+1}^{\mathrm{true}} - q_t^{\mathrm{true}} .
\end{equation}
The zero offset is removed entirely from the action. Under these idealized assumptions, even if the encoder zeros of A and C differ by 4$^\circ$, the recorded joint increments are identical as long as both execute the same true joint motion. Whether joint zero offsets contaminate training data therefore depends first on the state and action representations.

\subsection{Predicting Absolute Joint Targets Exposes the Model to Offsets}

\begin{figure}[tbp]
  \centering
  \includegraphics[width=0.62\linewidth]{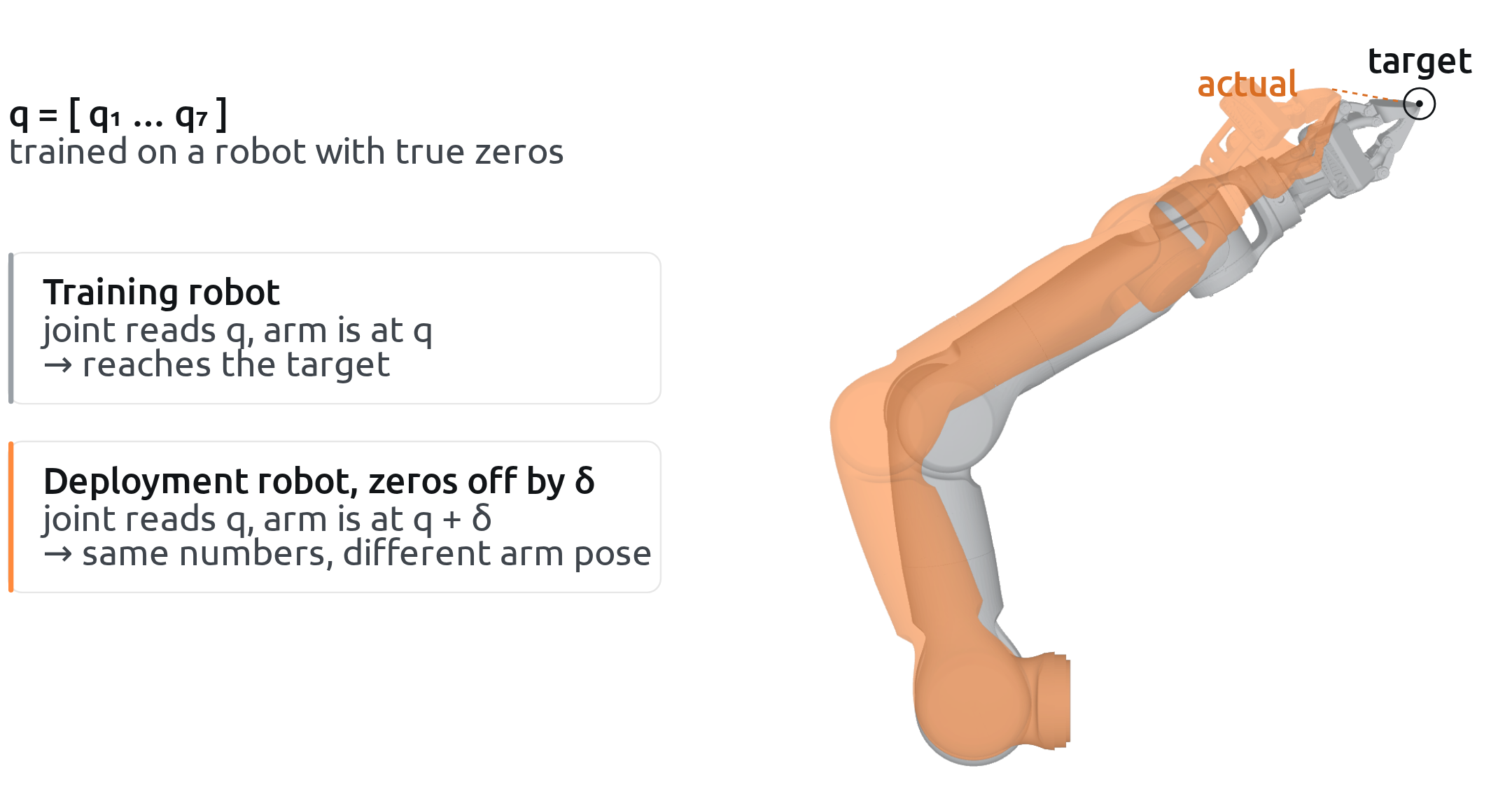}
  \caption{\textbf{The same command lands somewhere else on a miscalibrated robot.} A policy outputs joint angles; the robot that executes them has its own zero offsets, so the end effector misses the target.}
  \label{fig:app_inference}
\end{figure}

\textbf{Suppose the model predicts absolute joint positions mainly from images:}
\begin{equation}
  \pi(I_t) \rightarrow q_{t+1}^{\mathrm{target}} .
\end{equation}
For the same visual target, A provides the label 30$^\circ$ and C provides 34$^\circ$, so one input maps to two different supervision targets. If a deterministic policy is trained with MSE, then under the idealized condition that inputs are identical and the model cannot identify the data source, the optimal prediction is the conditional mean of the labels,
\begin{equation}
  \pi^*(I) = \mathbb{E}[a \mid I] .
\end{equation}
When different robots provide conflicting action labels for the same input, adding offset data can degrade performance on A (Figure~\ref{fig:app_inference}). The problem is not specific to MSE. A diffusion policy~\citep{chi2023diffusion} may learn a multimodal action distribution, generating actions near either 30$^\circ$ or 34$^\circ$ for the same image. Diffusion can represent such a distribution, but it does not know that the deployment robot is A; without enough conditioning information to distinguish robots, sampled actions may still be wrong for A. Even if the offsets of different C machines point in different directions and average to nearly zero, performance can still degrade: the mean bias cancels, but the conditional variance of the action labels grows. For tasks requiring precise contact, this dispersed action distribution can substantially reduce success.

\textbf{Suppose instead the model predicts absolute joint positions mainly from the joint state:}
\begin{equation}
  \pi(I_t, q_t^{\mathrm{obs}}) \rightarrow q_{t+1}^{\mathrm{target}} .
\end{equation}
Continuing the example, A's current state is 30$^\circ$ and C's is 34$^\circ$, and the image indicates that the relevant joint should increase by 5$^\circ$. The data from the two robots are
\begin{equation}
  A:\ (I, 30^\circ) \rightarrow 35^\circ, \qquad C:\ (I, 34^\circ) \rightarrow 39^\circ .
\end{equation}
States and actions still differ, but they obey exactly the same rule,
\begin{equation}
  a_t = q_t^{\mathrm{obs}} + \Delta q_t .
\end{equation}
A model that learns to predict the required joint increment from vision and the current state explains both datasets. Deployed on A, it sees 30$^\circ$ and naturally outputs 35$^\circ$. In this example, C's data are biased in absolute joint coordinates but introduce no contradictory control law. They may even act as beneficial augmentation: the model sees different joint-state values but must learn the same motion rule, which, for a network trained on limited data, can reduce its dependence on specific absolute joint values.

\textbf{Common imitation-learning models condition on both images and joint state, so their behavior lies between these two cases. In our observations, the more tokens the image occupies, the more strongly the model is driven by the image. Consequently, predicting absolute joint targets degrades performance under zero offsets.}

\textbf{To prevent offsets from contaminating the action, we use delta joint actions in this study, so that the model is affected by the offset only through the state.}

\clearpage
\section{Robot Models Included in the Study}
\label{app:robots}

Figure~\ref{fig:robot_inventory} lists the 25 robot models used in the embodiment study, grouped by the local model catalog into wheeled humanoids, mobile dual-arm systems, and tabletop dual-arm systems. Brand and model labels identify the corresponding visual models.

\begingroup
\setlength{\parskip}{2pt}
\newcommand{\robotcard}[3]{%
  \begin{minipage}[t][82pt][t]{0.19\linewidth}
    \centering
    \includegraphics[width=\linewidth,height=62pt,keepaspectratio]{figs/robots/#1.png}\\[2pt]
    {\sffamily\fontsize{7.4}{8.7}\selectfont #2\\\bfseries #3\par}%
  \end{minipage}%
}
{\sffamily\bfseries\color{reportorange}Wheeled humanoids}\hfill{\small\color{gray}15 models}\par\smallskip
\noindent\hfill%
\robotcard{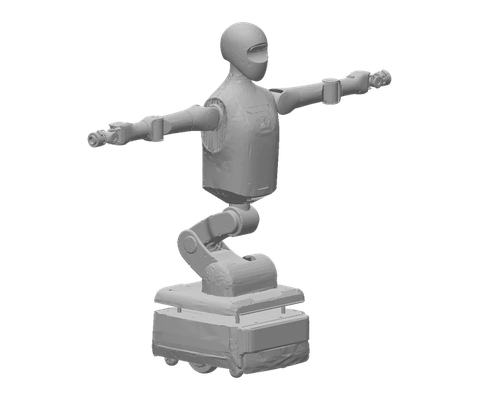}{X-Humanoid}{Tianyi 2.0}\hfill%
\robotcard{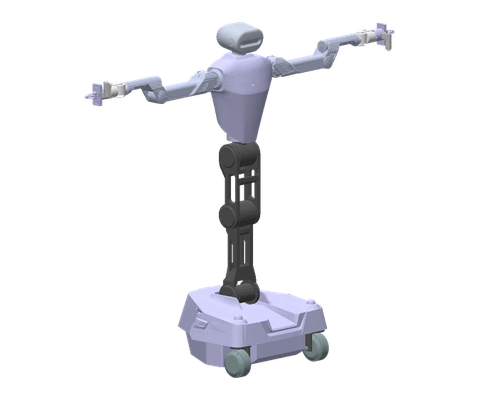}{Galaxea}{R1}\hfill%
\robotcard{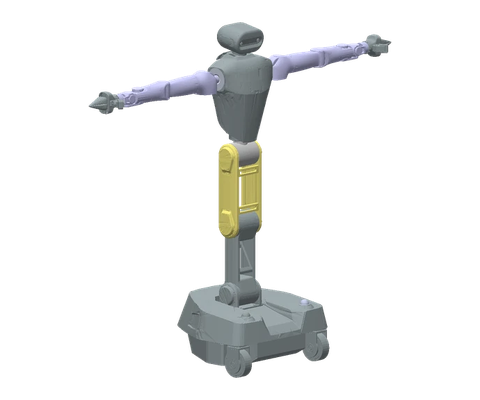}{Galaxea}{R1 Pro}\hfill%
\robotcard{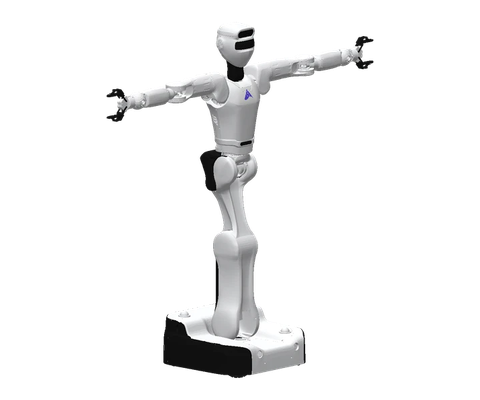}{Astribot}{S1}\hfill%
\robotcard{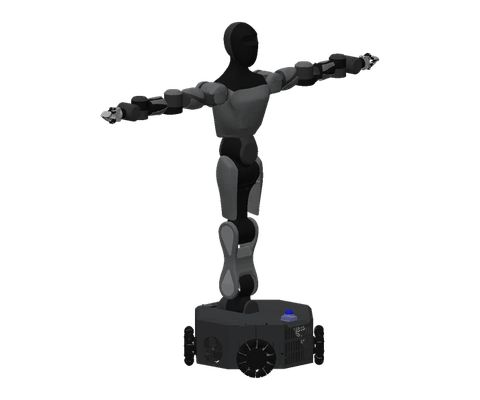}{Spirit AI}{MOZ1}\hfill\mbox{}\par
\noindent\hfill%
\robotcard{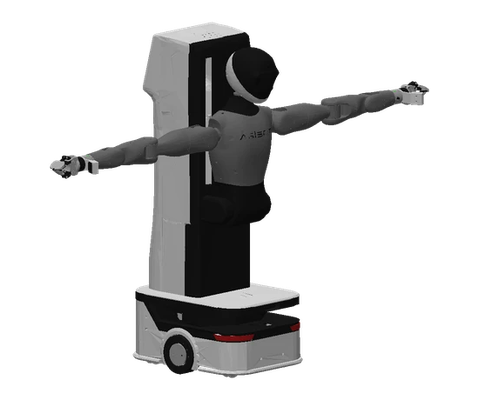}{AgiBot}{G1}\hfill%
\robotcard{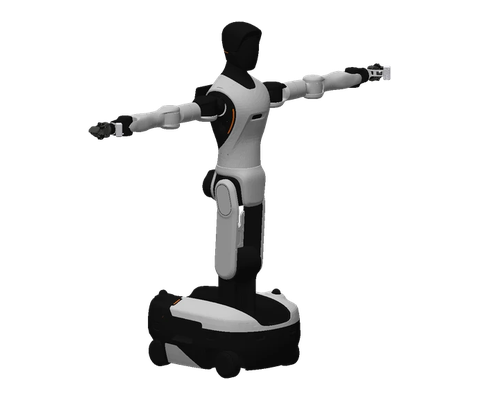}{AgiBot}{G2}\hfill%
\robotcard{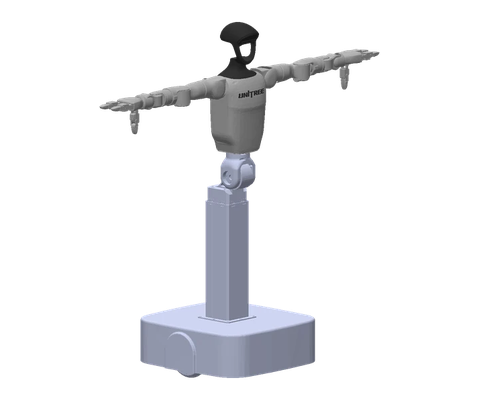}{Unitree}{G1-D}\hfill%
\robotcard{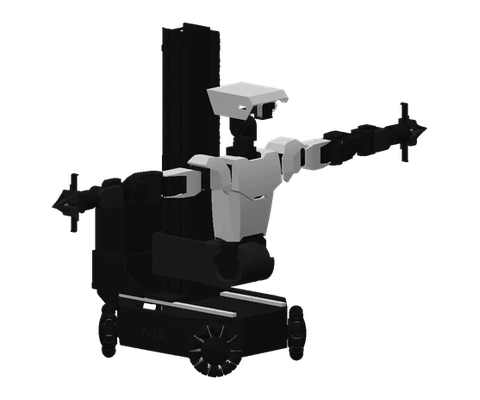}{ARX}{X7S}\hfill%
\robotcard{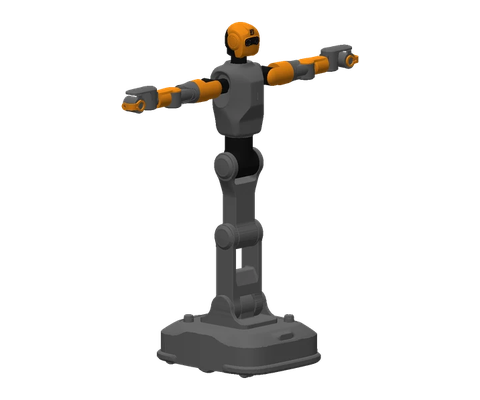}{DexForce}{W1}\hfill\mbox{}\par
\noindent\hfill%
\robotcard{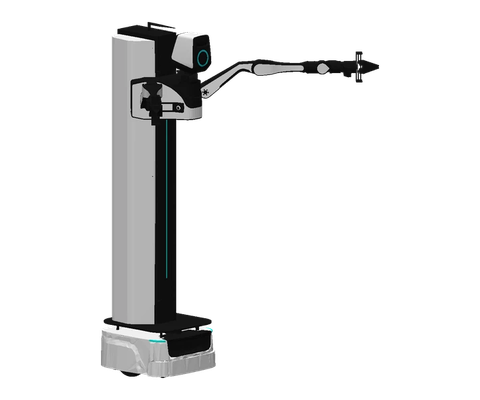}{Airbot}{MMK2}\hfill%
\robotcard{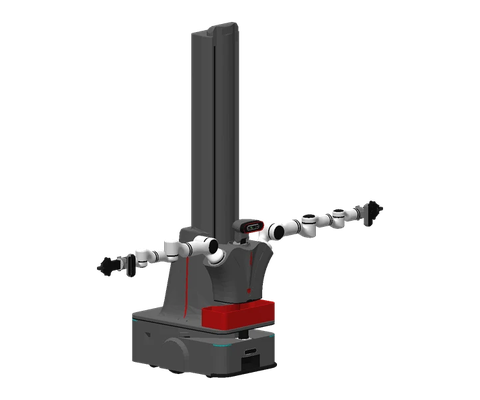}{RealMan}{AIDAL}\hfill%
\robotcard{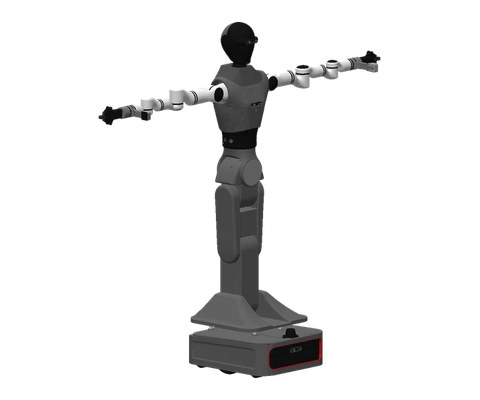}{RealMan}{RS-01}\hfill%
\robotcard{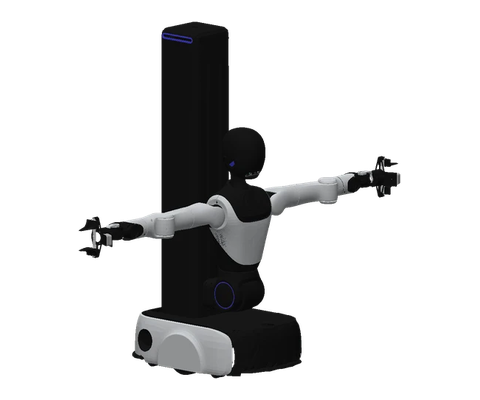}{Zerith}{H1}\hfill%
\robotcard{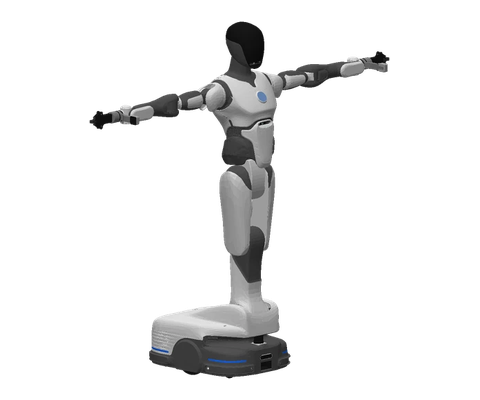}{Ai2 Robotics}{Bot2}\hfill\mbox{}\par
\smallskip
{\sffamily\bfseries\color{reportorange}Mobile dual-arm systems}\hfill{\small\color{gray}5 models}\par\smallskip
\noindent\hfill%
\robotcard{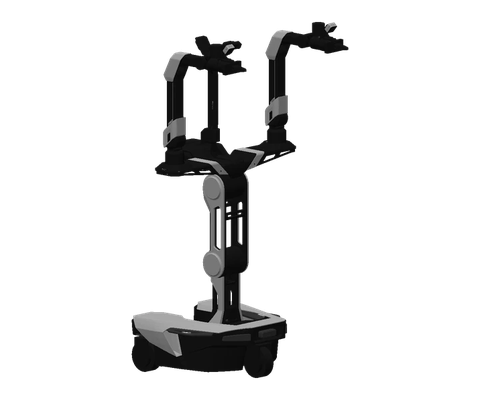}{Galaxea}{R1 Lite}\hfill%
\robotcard{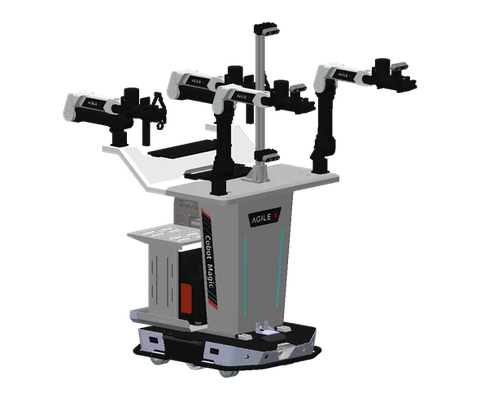}{AgileX}{Cobot Magic}\hfill%
\robotcard{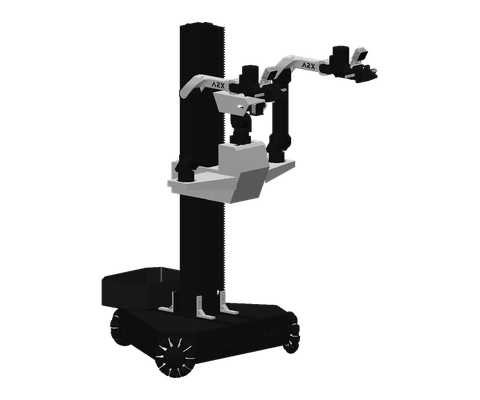}{ARX}{LIFT}\hfill%
\robotcard{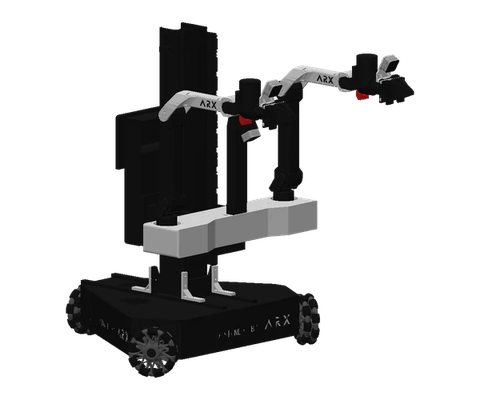}{ARX}{LIFT2S}\hfill%
\robotcard{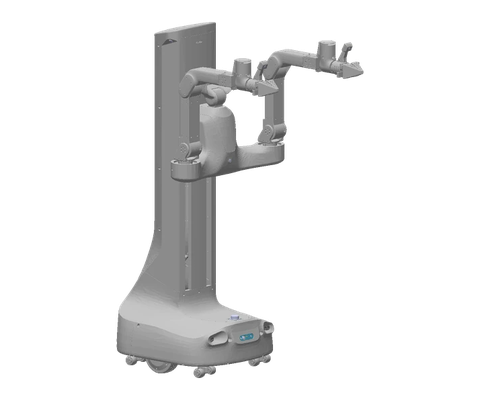}{XSquare Robot}{Quanta X1 Pro}\hfill\mbox{}\par
\smallskip
{\sffamily\bfseries\color{reportorange}Tabletop dual-arm systems}\hfill{\small\color{gray}5 models}\par\smallskip
\noindent\hfill%
\robotcard{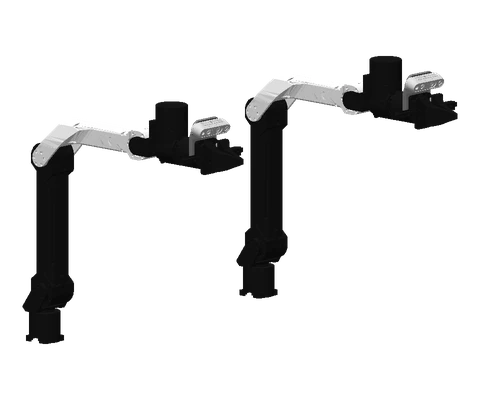}{ARX}{Dual X5}\hfill%
\robotcard{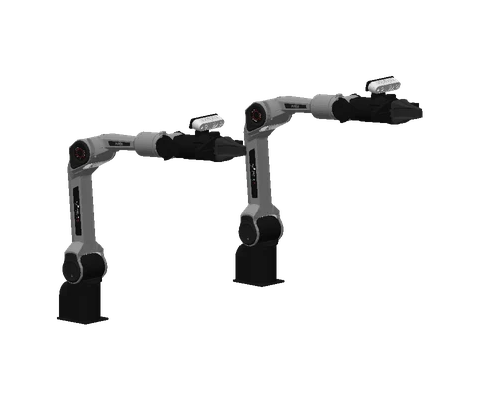}{AgileX}{Dual Piper}\hfill%
\robotcard{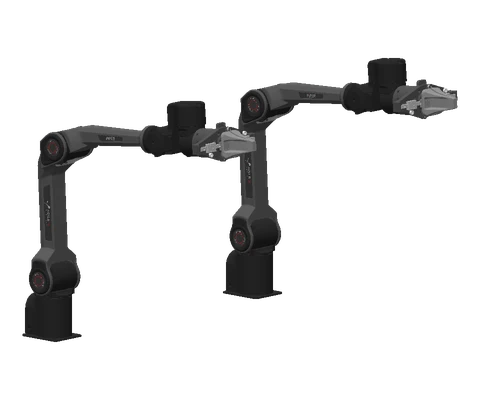}{AgileX}{Dual PiperX}\hfill%
\robotcard{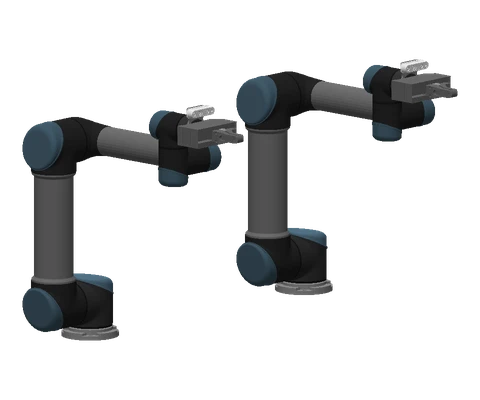}{Universal Robots}{Dual UR5}\hfill%
\robotcard{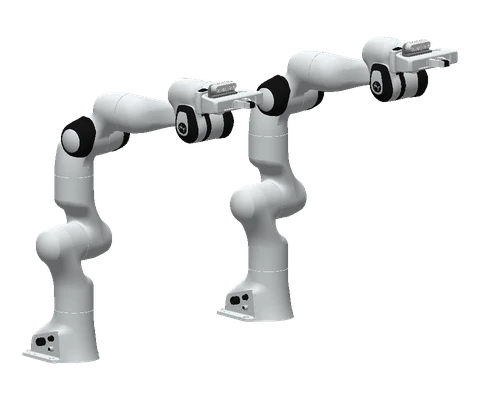}{Franka Robotics}{Dual Franka Panda}\hfill\mbox{}\par
\smallskip
\captionof{figure}{\textbf{The 25 robot models in the embodiment study.} Renderings use the local visual-model assets and their display poses. Robot sizes are normalized independently for readability and are not shown to physical scale.}
\label{fig:robot_inventory}
\endgroup

\end{document}